\pdfoutput=1

\documentclass[11pt]{article}

\usepackage[]{acl}

\usepackage{times}
\usepackage{latexsym}

\usepackage[T1]{fontenc}

\usepackage[utf8]{inputenc}

\usepackage{microtype}

\usepackage{inconsolata}

\usepackage{array}
\usepackage{bbm}
\usepackage{graphicx}
\usepackage{makecell}
\usepackage{xcolor}
\usepackage{soul}
\newcommand{\hlc}[2][yellow]{{%
		\colorlet{foo}{#1}%
		\sethlcolor{foo}\hl{#2}}%
}

\newcolumntype{P}[1]{>{\centering\arraybackslash}p{#1}}

\usepackage{amsmath,amsfonts,bm} 
\usepackage{pifont} 
\usepackage{graphicx,subfigure,epsfig,fancybox} 
\usepackage{float}
\usepackage{color} 
\usepackage{multirow}
\usepackage{natbib}

\usepackage{tikz}
\usetikzlibrary{shapes.geometric, positioning, calc}
\usetikzlibrary{arrows,shapes,calc}
\usetikzlibrary{trees,positioning,mindmap,shadows,fit}
\usetikzlibrary{decorations.pathreplacing}
\usetikzlibrary{tikzmark} 
\usetikzlibrary{intersections} 

\newcommand{\revised}[1]{\textcolor{blue}{}}

\usepackage{adjustbox}
\usepackage{array}
\usepackage{booktabs}

\newcolumntype{R}[2]{%
    >{\adjustbox{angle=#1,lap=\width-(#2)}\bgroup}%
    l%
    <{\egroup}%
}

\usepackage{setspace}

\usepackage{xspace}
\newcommand{\outputr}{\textsc{XY$^*$\textrightarrow R}\xspace}
\newcommand{\outputy}{\textsc{X\textrightarrow Y}\xspace}
\newcommand{\outputry}{\textsc{X\textrightarrow RY}\xspace}

\usepackage[capitalize]{cleveref}

\title{Benchmarking Large Language Models on Answering and Explaining Challenging Medical Questions}

\author{
Hanjie Chen$^{*1}$ \quad 
Zhouxiang Fang$^{*2}$ \quad
Yash Singla$^{2}$ \quad
Mark Dredze$^{2}$ \\
$^{1}$ Rice University \quad $^{2}$ Johns Hopkins University\\
\texttt{hanjie@rice.edu} \quad \texttt{\{zfang27, ysingla1\}@jh.edu} \quad \texttt{mdredze@cs.jhu.edu}
}

\begin{document}
\maketitle
\def\thefootnote{*}\makeatletter\def\Hy@Warning#1{}\makeatother\footnotetext{
Equal contribution, work done while Hanjie Chen was a postdoc at JHU}
\def\thefootnote{\arabic{footnote}}
\begin{abstract}
LLMs have demonstrated impressive performance in answering medical questions, such as achieving passing scores on medical licensing examinations. 
However, medical board exams or general clinical questions do not capture the complexity of realistic clinical cases. 
Moreover, the lack of reference explanations means we cannot easily evaluate the reasoning of model decisions, a crucial component of supporting doctors in making complex medical decisions. 
To address these challenges, we construct two new datasets: JAMA Clinical Challenge and Medbullets.\footnote{Datasets and code are available at \url{https://github.com/HanjieChen/ChallengeClinicalQA}.}
JAMA Clinical Challenge consists of questions based on challenging clinical cases, while Medbullets comprises simulated clinical questions. 
Both datasets are structured as multiple-choice question-answering tasks, accompanied by expert-written explanations. 
We evaluate seven LLMs on the two datasets using various prompts. 
Experiments demonstrate that our datasets are harder than previous benchmarks. 
In-depth automatic and human evaluations of model-generated explanations provide insights into the promise and deficiency of LLMs for explainable medical QA. 
\end{abstract}

\section{Introduction}
\label{sec:intro}
Medical question answering (QA) involves synthesizing relevant information and knowledge and then reasoning about how to apply them to the current situation.
These characteristics, presented in many large language models (LLMs) \citep{brown2020language, almazrouei2023falcon, chowdhery2023palm, touvron2023llama}, have been demonstrated in the case of answering medical questions, with some claims that LLMs can achieve passing scores on medical board exams \citep{singhal2022large, nori2023can}. 
These results have been achieved with both domain-specific LLMs, such as Med-PaLM \citep{singhal2022large} and MEDITRON \citep{chen2023meditron}, as well as general-purpose LLMs, such as GPT-4 \citep{openai2023gpt}, on the United States Medical Licensing Examination (USMLE) questions \citep{nori2023capabilities}.

While high accuracy on medical licensing exams sounds impressive, it does not mean that these models can answer complex medical questions or assist doctors with clinical decisions. 
Board exam questions \citep{jin2021disease, pal2022medmcqa} and general medical questions \citep{abacha2019bridging, abacha2017overview} rely on textbook knowledge, a task for which LLMs are well suited since they learn primarily from medical texts. 
In contrast, doctors need support with clinical presentations that differ from textbook definitions and require blending established knowledge with clinical experience \citep{harris2023large, kanjee2023accuracy}. 
Additionally, doctors seek explanations, beyond predictions, that elucidate reasons for a decision to better understand complex clinical cases \citep{panigutti2022understanding}. 
Models should be evaluated on their ability to correctly explain complex medical decisions in addition to making them.

\begin{figure*}[t]
  \centering
  \includegraphics[width=0.95\textwidth]{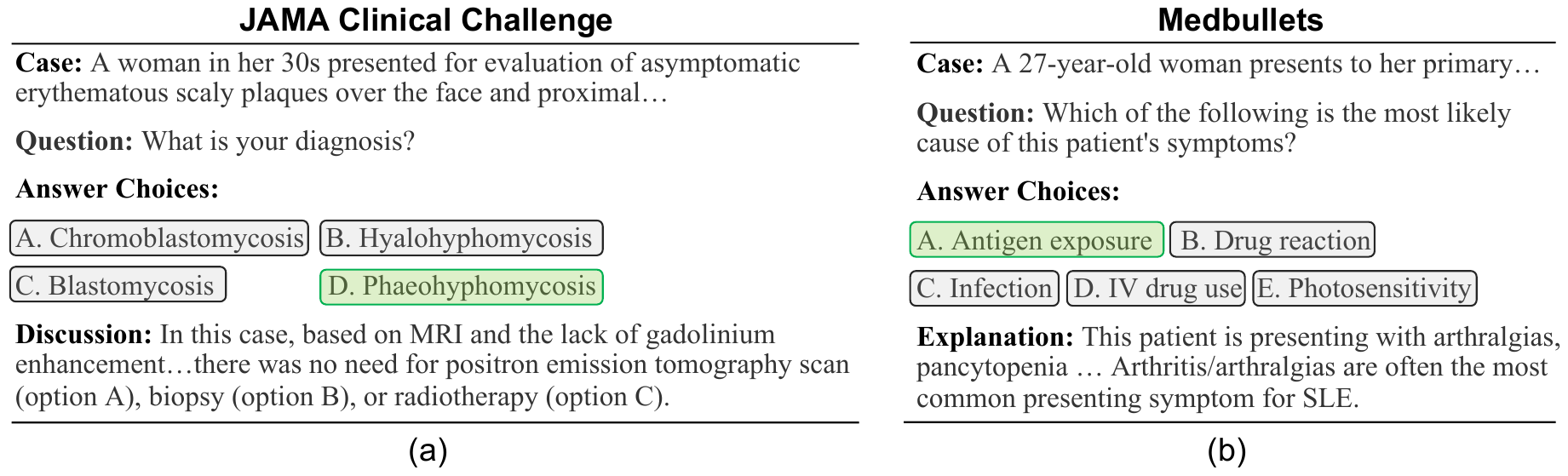}
    \caption{\label{fig:illustration}
  Examples from the JAMA Clinical Challenge and Medbullets datasets respectively. Each question is paired with 4/5 answer choices and a discussion/explanation. The correct answers are highlighted in green boxes. 
  }
\end{figure*}

Moving the evaluation of the medical capabilities of LLMs toward more realistic and challenging clinical settings requires supportive benchmarks. 
There have been some attempts in this regard, but they are limited in scope and the datasets are generally small \citep{strong2023chatbot, kanjee2023accuracy, shea2023use, jin2024hidden}. 
Some benchmarks evaluate model capabilities in understanding clinical texts or following clinical instructions, without necessitating complex reasoning over synthesized information and knowledge for clinical decisions \citep{parmar2023longbox, fleming2023medalign, he2023medeval}. 
While recent benchmarks include questions derived from real-world conversations or closely aligned with real-world practice \citep{manes2024k, liu2024large, wang2024assessing}, they do not provide explanations. 
Other benchmarks include explanations for medical exam questions \citep{pal2022medmcqa, kim-etal-2024-medexqa}. 
To the best of our knowledge, our datasets are the first to feature challenging real-world clinical questions accompanied by high-quality, expert-written explanations.

We address this evaluation gap through the construction of two datasets: JAMA Clinical Challenge and Medbullets (\autoref{fig:illustration}). 
Both datasets contain high-quality explanations written by human experts---a distinctive feature lacking in previous benchmarks.

JAMA Clinical Challenge consists of 1524 clinical cases collected from the JAMA Network Clinical Challenge archive. 
These articles are summaries of actual challenging clinical cases framed as challenging questions. 
Each article contains a long case description followed by a question, four answer choices, and a discussion (explanation)\footnote{For consistency, we will use ``explanation'' in this paper.} explaining the correct (highlighted in green) and incorrect answers. 
The questions cover a wide range of medical topics. 
We expect these questions to be especially challenging given their length and complexity.

Medbullets comprises 308 USMLE Step 2\&3 style questions collected from open-access tweets on X (formerly Twitter) since April 2022. 
The difficulty is comparable to that of Step 2\&3 exams, which emulate common clinical scenarios but are shorter and (perhaps) less challenging than JAMA.
Each question is paired with a case description, five answer choices, and an explanation of correct and incorrect answers. 
Although existing benchmarks have included the same type of questions (e.g., MedQA \citep{jin2021disease}), our data contains answer explanations and is more challenging due to its recency, which has led to a significant drop in model performance (\S\ref{sec:pred_acc}).
 
We evaluate seven LLMs, spanning closed-source and open-source, general-purpose and domain-specific models: GPT-3.5 \citep{ouyang2022training}, GPT-4 \citep{openai2023gpt}, PaLM 2 \citep{anil2023palm}, Llama 2 \citep{touvron2023llama}, Llama 3 \citep{meta2024llama}, MedAlpaca \citep{han2023medalpaca}, and Meerkat \citep{kim2024small}. 
We find our datasets to be harder (lower accuracy) compared to previous benchmarks, highlighting the challenges posed by the new tasks. 
In-context learning and prompting strategies yield marginal improvements, leaving potential areas for improvement in future work. 
In-depth automatic and human evaluations of model-generated explanations highlight the limitations of LLMs in explaining complex medical decision-making. 
Moreover, the weak correlation between human and automatic scores underscores the necessity of developing metrics that can support future research on explainable medical QA. 
Overall, these datasets represent more challenging goals for model predictions and a new challenge in producing meaningful explanations for medical decisions.
\section{Datasets}
\label{sec:datasets}
We present two medical QA datasets with explanations, both in English. 

\subsection{JAMA Clinical Challenge}
\label{sec:jama}
The Journal of the American Medical Association (JAMA) includes a Clinical Challenge feature, which presents challenging real-world clinical cases from a range of medical domains, such as ophthalmology, dermatology, etc. 
These cases are purposely selected based on their difficulty and unusual presentation.
\autoref{fig:illustration} (a) shows an example case of a single-patient with a specific disease or condition. 
Each case has a question about diagnosis (``What is your diagnosis?'') or decision (``What would you do next?'').
Four answer choices follow, with the correct option highlighted in the green box (\autoref{fig:illustration}).
A reference explanation describes why the correct answer is preferred over others.

\paragraph{Data Statistics}
We collected 1524 examples from
the JAMA Network Clinical Challenge\footnote{\url{https://jamanetwork.com/collections/44038/clinical-challenge}} archive, spanning the past decade (July 2013 - October 2023) and 13 medical domains. 
For this paper we excluded images to focus on text LLMs. 
\autoref{tab:med_stats} shows data statistics, and more details across domains are reported in \autoref{tab:jama_stats}, where ``General'' means no specific field was provided.
We calculate the average and maximum lengths (in alphanumeric tokens) of inputs (Description+Question+Options) and explanations respectively. 
Ophthalmology and Dermatology contain many more examples than other domains, while Psychiatry has only five instances, which are much longer than average. 
The explanation length is longer in Diagnostic due to questions about interpretations for diagnostic results, leading to relatively long background introductions and analyses for each potential diagnosis.

\subsection{Medbullets}
\label{sec:medbullets}
Medbullets is an online platform that provides resources for medical study. 
We focus on Medbullets Step 2/3\footnote{\url{https://step2.medbullets.com/}} which serves USMLE Step 2\&3\footnote{\url{https://www.usmle.org/step-exams/step-2-ck}} type questions. 
The difficulty level of these questions surpasses that of Step 1 questions, which primarily rely on textbook knowledge. 
Solving Step 2/3 questions demands the application of medical knowledge and clinical reasoning. 
In this paper, Medbullets refers to Medbullets Step 2/3.
\autoref{fig:illustration} (b) shows an example from Medbullets, comprising a case description, a question, five answer choices -- with the correct one highlighted in the green box -- and an explanation that explains each option. 

\begin{table}[t]
    \centering
    \small
    \begin{tabular}{cccc}
    \toprule
     Dataset & \textit{\#} & $A_{in}/M_{in}$ & $A_{exp}/M_{exp}$ \\
     \midrule
     JAMA & 1524 & 371/779 &  632/1369 \\
     \midrule
     Medbullets &  308 &  163/257&  413/ 756 \\
     \midrule
     MedQA & 1273 &  137/529&  -  \\
    \bottomrule
    \end{tabular}
    \caption{Statistics of JAMA Clinical Challenge (JAMA), Medbullets, and MedQA, where \textit{\#} counts the number of test examples, $A_{in}/M_{in}$ means the average/maximum length of inputs, and $A_{exp}/M_{exp}$ is the average/maximum length of explanations.}
    \label{tab:med_stats}
\end{table}
\begin{figure*}[t]
  \centering
  \includegraphics[width=1\textwidth]{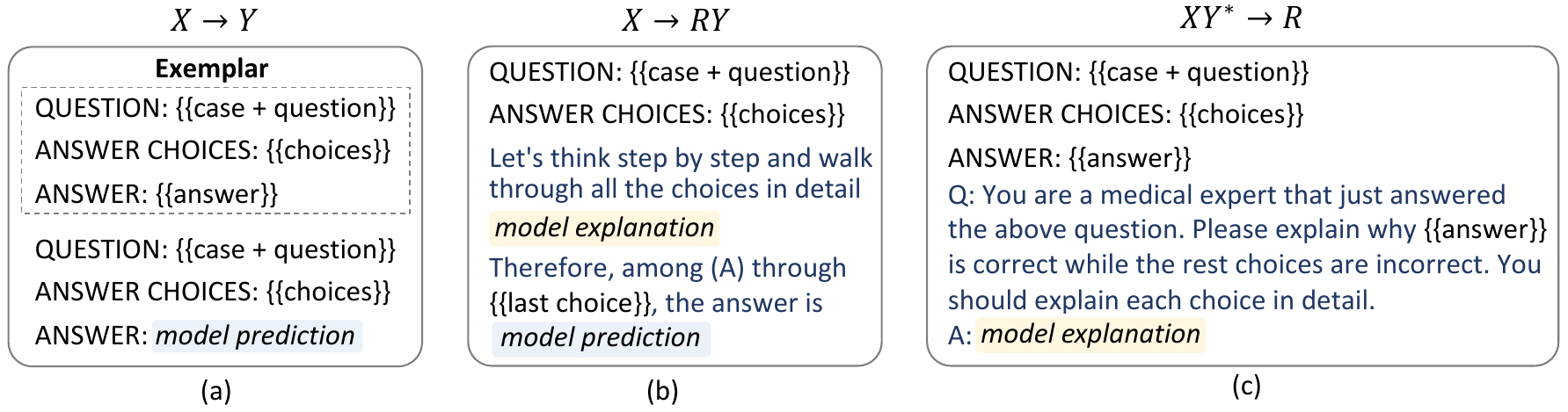}
    \caption{\label{fig:prompt}
  Prompt templates. The contents in \{\{\}\} are replaced with specific elements. Model predictions and explanations are highlighted in blue and yellow respectively.
  }
\end{figure*}

\paragraph{Data Statistics}
Medbullets posted links to questions on X\footnote{\url{https://x.com/medbullets}}, which we used to collect 308 examples that were publicly available from April 2022 to December 2023. 
As before we excluded images. 
\autoref{tab:med_stats} shows the statistics of Medbullets, where the average/maximum lengths of inputs and explanations are shorter than those of JAMA.

MedQA \citep{jin2021disease} also included questions from Medbullets.
The test set includes 679 Step 1 questions and 594 Step 2/3 questions, obtained in March 2021. 
We exclusively consider Step 2/3 questions, leading to a larger average input length compared to MedQA (\autoref{tab:med_stats}).
Our data is both more recent -- none of our questions appear in MedQA -- and contains explanations.
MedQA questions have 5 answer options (MedQA-5) while another version (MedQA-4) has only 4 options, where a randomly selected incorrect option was dropped. 
We also created both 5-option and 4-option versions of Medbullets, denoted as \textbf{Medbullets-5} and \textbf{Medbullets-4} respectively, for comparison. 

More details on data sources and collection, as well as examinations of data robustness and contamination, can be found in Appendix \ref{sec:add_data}. 
Empirical studies show that our datasets do not have robustness or contamination issues \citep{alzahrani-etal-2024-benchmarks, sainz-etal-2023-nlp}
\section{Experimental Setup}
\label{sec:setup}
We evaluate a range of LLMs on the datasets to determine if they are more challenging than previous benchmarks. We also evaluate the ability of models to produce explanations.
We start by describing the LLMs in our experiments (\S\ref{sec:models}), followed by prompting strategies (\S\ref{sec:prompts}), and evaluation metrics (\S\ref{sec:metrics}). 
More details are in Appendix \ref{sec:add_setup}.

\subsection{Models}
\label{sec:models}
We investigate a range of LLMs: GPT-3.5 \cite[\texttt{gpt-3.5-turbo-0613}]{ouyang2022training}, GPT-4 \cite[\texttt{gpt-4-0613}]{openai2023gpt}, PaLM 2 \cite[\texttt{chat-bison-001}]{anil2023palm}, Llama 2 \cite[\texttt{Llama-2-70b-chat}]{touvron2023llama}, Llama 3 \cite[\texttt{Llama-3-70b-chat}]{meta2024llama}, MedAlpaca \cite[\texttt{medalpaca-13b}]{han2023medalpaca}, and Meerkat \cite[\texttt{meerkat-7b-v1.0}]{kim2024small}. 
GPT-3.5, GPT-4, and PaLM 2 are closed-source, general-purpose models. 
Llama 2 and Llama 3 are open-source, general-purpose models. 
MedAlpaca and Meerkat are open-source, domain-specific models tailored to the medical domain. 
We chose these two medical models due to their superiority over other candidate models in our pilot experiments (\S\ref{sec:add_setup_model}) and in previous studies \citep{kim2024small, zhang2023alpacare}.

\subsection{Prompting Strategies}
\label{sec:prompts}
We apply three different prompting strategies to generate answers and/or explanations.
We represent the input as $X$ for the description, question and answer options, $Y$ for the answer, and $R$ for the explanation (rationale). 
Our three strategies (\autoref{fig:prompt}) are:
\begin{itemize}
\item \outputy: ask the model to answer the question.
\item \outputry: ask the model to engage in step-by-step reasoning first, and then answer the question.
This strategy is based on \textit{chain-of-thought} (CoT) prompting, which has improved LLMs' prediction accuracy across various reasoning tasks \citep{wei2022chain}. 
We follow the two-stage prompting scheme in \citet{lievin2022can, nori2023capabilities}, as illustrated in \autoref{fig:prompt} (b), where the CoT cue ``\textit{Let's think step by step and walk through all the choices in detail.}'' is adapted from \citet{kojima2022large}. 
In the first stage, the model produces its explanation (highlighted in yellow). 
We then provide the entire generation to the second-stage and ask the model to provide a prediction (highlighted in blue). 
\item \outputr: given the input and the correct answer (Y$^*$), ask the model to explain why this is the best answer over other options. This prompting produces an explanation only.
\end{itemize}
More details of specific prompts for each LLM are shown in \Crefrange{tab:prompt_detail_1}{tab:prompt_detail_3}.

\paragraph{Few-Shot Prompting}
LLMs have demonstrated their capability for in-context learning by utilizing exemplars, enabling them to fastly adapt to new tasks through few-shot prompting \citep{brown2020language, min-etal-2022-rethinking, chen-etal-2022-meta}. 
We modify the prompt to include few-shot examples, shown in the dashed box in \autoref{fig:prompt} (a). 
Each exemplar uses the same template of the zero-shot prompting above, with the model output replaced by the gold answer or explanation. 

\subsection{Evaluation Metrics}
\label{sec:metrics}
We use accuracy (prediction compared to ground truth) to evaluate the predictions of each model. 
We explore several methods to evaluate different aspects of model-generated explanations: ROUGE-L \citep{lin2004rouge}, BERTScore \citep{zhang2019bertscore}, BLEURT \citep{sellam-etal-2020-bleurt}, two variants of BARTScore---BARTScore+ and BARTScore++ \citep{yuan2021bartscore}, three metrics of CTC \citep{deng-etal-2021-compression}---Consistency, Relevance, and Preservation, and three metrics of G-Eval \citep{liu-etal-2023-g}---Coherence, Consistency, and Relevance. 
ROUGE-L computes the surface-form similarity between model-generated explanations and reference (gold) explanations (\S\ref{sec:datasets}). 
BERTScore, BLEURT, and BARTScore+(+) measure semantic similarity using pre-trained BERT \citep{devlin-etal-2019-bert}, fine-tuned BERT, and fine-tuned BART \citep{lewis-etal-2020-bart} models respectively. 
CTC metrics evaluate the information alignment of model-generated explanations w.r.t. reference explanations or inputs. 
G-Eval utilizes an LLM (e.g., GPT-4) as the backbone model to score the quality of model-generated explanations in different aspects. 
More details are in Appendix \ref{sec:add_metric}. 
\section{Results and Discussion}
\label{sec:results}
We measure the prediction accuracy of the seven LLMs on MedQA and our two datasets with the \outputy prompt (\S\ref{sec:pred_acc}), few-shot prompts (\outputy, \S\ref{sec:few_shot}) and CoT reasoning (\outputry, \S\ref{sec:cot}).
We evaluate model-generated explanations under \outputr prompting via both automatic and human evaluations (\S\ref{sec:exp_eval}).

\begin{table*}[t]
    \centering
    \small
    \begin{tabular}{P{2cm}@{\hskip 0.2cm}P{3cm}@{\hskip 0.2cm}P{1.2cm}@{\hskip 0.2cm}P{1cm}@{\hskip 0.2cm}P{1cm}@{\hskip 0.2cm}P{1.2cm}@{\hskip 0.2cm}P{1.2cm}@{\hskip 0.2cm}P{1.5cm}@{\hskip 0.2cm}P{1.2cm}@{\hskip 0.2cm}}
    \toprule
     \multirow{2}{*}{Prompting} &\multirow{2}{*}{Dataset} & \multicolumn{7}{c}{Model} \\
     \cmidrule(lr){3-9}
     & & GPT-3.5 & GPT-4 & PaLM 2 & Llama 2 & Llama 3 & MedAlpaca & Meerkat \\
     \midrule
     \multirow{9}{*}{\outputy} & MedQA-4 & 54.67 & 78.63 & 52.95 & 39.51 & 76.74 & 45.16 & 64.18 \\
     \rule{0pt}{2ex}
     & \textcolor{gray}{MedQA-4 (Step 1)} & \textcolor{gray}{54.79} & \textcolor{gray}{77.76} & \textcolor{gray}{53.31} & \textcolor{gray}{40.35} & \textcolor{gray}{75.55} & \textcolor{gray}{41.38} & \textcolor{gray}{62.74} \\
     \rule{0pt}{2ex}
     & \textcolor{gray}{MedQA-4 (Step 2/3)} & \textcolor{gray}{54.55} & \textcolor{gray}{79.63} & \textcolor{gray}{52.53} & \textcolor{gray}{38.55} & \textcolor{gray}{78.11} & \textcolor{gray}{49.49} & \textcolor{gray}{65.82} \\
     \rule{0pt}{2ex}
     & MedQA-5 & 48.55 & 74.16 & 46.58 & 32.76 & 72.58 & 39.90 & 61.19 \\
     \rule{0pt}{2ex}
     & \textcolor{gray}{MedQA-5 (Step 1)} & \textcolor{gray}{48.01} & \textcolor{gray}{73.34} & \textcolor{gray}{45.51} & \textcolor{gray}{34.02} & \textcolor{gray}{70.83} & \textcolor{gray}{34.75} & \textcolor{gray}{58.76} \\
     \rule{0pt}{2ex}
     & \textcolor{gray}{MedQA-5 (Step 2/3)} & \textcolor{gray}{49.16} & \textcolor{gray}{75.08} & \textcolor{gray}{47.81} & \textcolor{gray}{31.31} & \textcolor{gray}{74.57} & \textcolor{gray}{45.79} & \textcolor{gray}{63.94} \\
     \rule{0pt}{2ex}
     & Medbullets-4 & 46.10 & 66.23 & 47.73 & 33.12 & 68.50 & 48.37 & 56.49 \\
     \rule{0pt}{2ex}
     & Medbullets-5 & 43.18 & 60.71 & 42.86 & 25.97 & 63.96 & 43.18 & 48.70 \\
     \rule{0pt}{2ex}
     & JAMA & 48.69 & 67.32 & 48.69 & 44.36 & 66.14 & 36.48 & 45.99 \\
     \midrule
     \multirow{5}{*}{\outputry} & MedQA-4 & \textbf{57.42} & \textbf{82.64} & \textbf{54.44} & \textbf{40.22} & \textbf{78.79} & 44.70 & \textbf{68.89} \\
     \rule{0pt}{2ex}
     & MedQA-5 & \textbf{55.30} & \textbf{79.42} & \textbf{48.70} & \textbf{36.61} & \textbf{75.26} & 37.86 & \textbf{64.33} \\
     \rule{0pt}{2ex}
     & Medbullets-4 & \textbf{50.97} & \textbf{68.83} & 47.73 & \textbf{35.06} & 67.21 & 47.07 & \textbf{56.81}\\
      \rule{0pt}{2ex}
     & Medbullets-5 & \textbf{47.40} & \textbf{63.31} & \textbf{43.51} & \textbf{27.92} & 62.66 & 42.53 & \textbf{49.35} \\
      \rule{0pt}{2ex}
     & JAMA & \textbf{50.13} & 67.13 & 47.83 & 40.88 & 64.17 & 35.69 & \textbf{49.86} \\
    \bottomrule
    \end{tabular}
    \caption{Prediction accuracy (\%) of the seven LLMs under zero-shot \outputy and \outputry promptings. JAMA: JAMA Clinical Challenge. For \outputy, we report the results for the Step 1 and Step 2/3 questions of MedQA-4/5 separately in \textcolor{gray}{gray}. The results in \outputry that are better than the corresponding results in \outputy are bolded.}
    \label{tab:pred_eval}
\end{table*}

\subsection{Performance Drop on New Tasks}
\label{sec:pred_acc}
\autoref{tab:pred_eval} shows the accuracy using zero-shot \outputy prompting on the 4-option and 5-option versions of Medbullets and MedQA. 
All models, except for MedAlpaca, exhibit lower prediction accuracy on Medbullets-4/5 compared to MedQA-4/5, showing the challenge of our datasets. 
GPT-4 (by far the best model overall) drops over 12\%, while the other models decrease a range of 5\% to 12\%. 
We compare JAMA Clinical Challenge (4 answer choices) to Medbullets-4 and MedQA-4, where GPT-3.5, GPT-4, PaLM 2, and Llama 3 on JAMA Clinical Challenge results align closely with their performance on Medbullets-4.
This similarity implies that these models face comparable difficulties when addressing real-world clinical problems as they do with simulated USMLE Step 2/3 questions, a surprising finding given the challenge of JAMA questions. 
The performance of MedAlpaca and Meerkat drops significantly on JAMA Clinical Challenge, indicating that this task is challenging for smaller language models (13B and 7B respectively), even with fine-tuning on medical data. 
Interestingly, Llama 2 performs comparably to GPT-3.5, PaLM 2, and Meerkat on JAMA Clinical Challenge but falls short on Medbullets and MedQA. 
We analyze model performance across medical domains of JAMA Clinical Challenge in Appendix \ref{sec:jama_domains}. 

\begin{figure*}[t]
  \centering
  \includegraphics[width=0.9\textwidth]{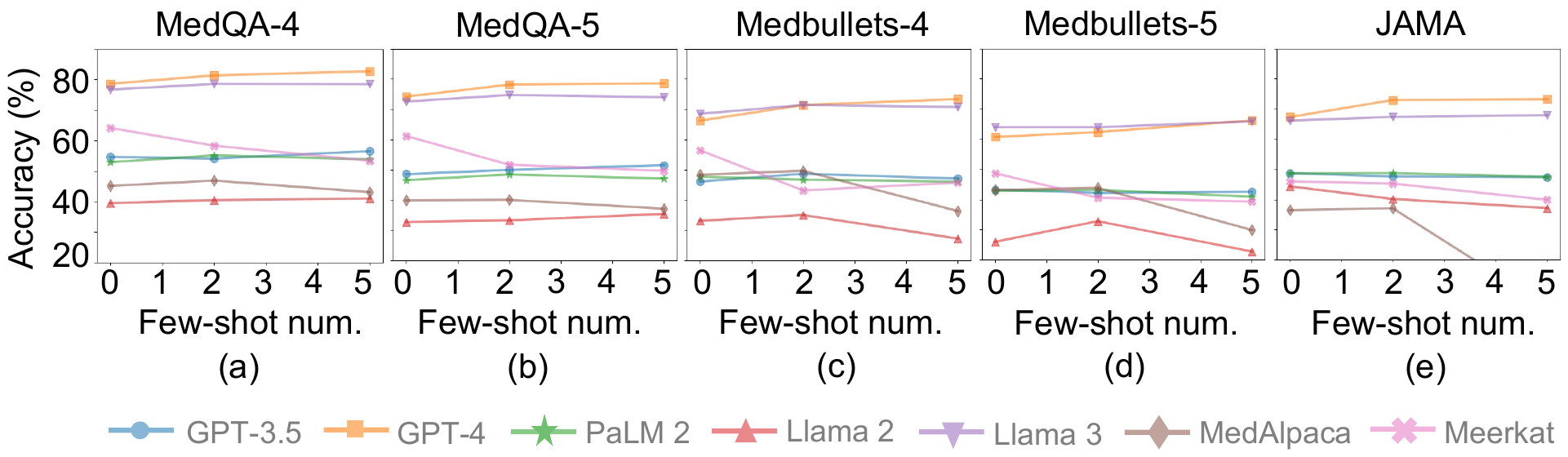}
    \caption{\label{fig:fewshot}
  In-context learning performance of the seven LLMs under few-shot \outputy prompting on different datasets.
  }
\end{figure*}

\paragraph{Is MedQA Easier Because It Includes Step 1 Questions?} 
MedQA contains Step 1, 2/3 questions. Since Step 1 questions should be easier, do LLMs do better on these questions? 
We compute the accuracy of the seven models on the Step 1 and Step 2/3 questions in MedQA-4/5 separately. 
The results are reported in \autoref{tab:pred_eval} in gray. 
All models, except for MedAlpaca, demonstrate similar performance in answering the two types of questions, meaning that the reason why MedQA is easier than Medbullets is not due to the inclusion of Step 1 questions. 
MedAlpaca's accuracy on MedQA Step 2/3 questions is much higher than on Step 1 questions and is comparable to its performance on Medbullets, which only contains Step 2/3 questions. 
This explains why its overall accuracy on MedQA (both 4- and 5-option versions) is lower than on Medbullets. 
Additionally, the other models perform better on Step 2/3 questions in MedQA than on Medbullets, suggesting that the newer questions in Medbullets are more challenging. 

\subsection{Does In-Context Learning Help?}
\label{sec:few_shot}
We follow \citet{nori2023capabilities} and apply few-shot \outputy prompting, illustrated in \autoref{fig:prompt} (a), to test the capabilities of LLMs in adapting to new tasks through in-context learning \citep{brown2020language}. 
Specifically, we adopt leave-one-out cross validation \citep{hastie2009elements}, where each instance is assessed using few-shot randomly sampled examples from the remaining dataset as demonstrations. 
\autoref{fig:fewshot} shows the prediction accuracy (\%) of the seven models under 0/2/5-shot \outputy promptings on different datasets. 
GPT-4 and Llama 3 benefit from in-context learning, exhibiting improved performance with an increased number of few-shot examples across all datasets. 
Nevertheless, it has minimal effect for GPT-3.5 and PaLM 2, whose performances are tied and do not show obvious differences under few-shot prompting. 
Llama 2 and MedAlpaca's performance remains stable on MedQA-4/5 but varies on Medbullets-4/5 and declines on JAMA Clinical Challenge when more examples are used. 
MedAlpaca can barely output reasonably under 5-shot prompting on JAMA due to its 512-token limit. 
Few-shot prompting generally hurts Meerkat, implying its inferior in-context learning ability. 
Overall, few-shot prompting does not enhance the adaptation of GPT-3.5, PaLM 2, Llama 2, MedAlpaca, and Meerkat to the new tasks. 
We leave it to future work to develop better learning or adaptation strategies to equip LLMs with the capabilities to tackle challenging medical QA. 

\begin{table*}[t]
    \centering
    \small
    \begin{tabular}{P{2cm}@{\hskip 0.2cm}P{3cm}@{\hskip 0.2cm}P{1.2cm}@{\hskip 0.2cm}P{1cm}@{\hskip 0.2cm}P{1cm}@{\hskip 0.2cm}P{1.2cm}@{\hskip 0.2cm}P{1.2cm}@{\hskip 0.2cm}P{1.5cm}@{\hskip 0.2cm}P{1.2cm}@{\hskip 0.2cm}}
    \toprule
     \multirow{2}{*}{Dataset} &\multirow{2}{*}{Metric} & \multicolumn{7}{c}{Model} \\
     \cmidrule(lr){3-9}
     & & GPT-3.5 & GPT-4 & PaLM 2 & Llama 2 & Llama 3 & MedAlpaca & Meerkat \\
     \midrule
     \multirow{6}{*}{Medbullets-5} & ROUGE-L & \textbf{0.3323} & \underline{0.3119} & 0.2995 & 0.3000 & 0.3118 & \textcolor{gray}{0.1146} & 0.3107 \\
     \rule{0pt}{2ex}
     & BERTScore & \textbf{0.6554} & \underline{0.6530} & 0.6300 & 0.6358 & 0.6450 & \textcolor{gray}{0.4658} & 0.6449 \\
     \rule{0pt}{2ex}
     & BLEURT & 0.3965  & \underline{0.3981}  & 0.3898  & 0.3787  & 0.3915  & \textcolor{gray}{0.3712}  & \textbf{0.3985}  \\
     \rule{0pt}{2ex}
     & CTC Relev. & 0.6961 & \underline{0.6965} & 0.6835 & 0.6898 & 0.6920 & \textcolor{gray}{0.6707} & \textbf{0.7057} \\
     \rule{0pt}{2ex}
     & CTC Presv. & \textbf{0.4255} & \underline{0.4251} & 0.4161 & 0.4222 & 0.4244 & \textcolor{gray}{0.3935} & 0.4243 \\
     \rule{0pt}{2ex}
     & CTC Consist. & 0.8291 & 0.8253 & \textcolor{gray}{0.8209} & 0.8294 & 0.8237 & \textbf{0.8724} & \underline{0.8406} \\
     \rule{0pt}{2ex}
     & G-Eval Relev. & 4.6449 & \underline{4.7712} & 3.9566 & 3.9943 & \textbf{4.9103} & \textcolor{gray}{1.9727} & 4.0233 \\
     \rule{0pt}{2ex}
     & G-Eval Coh. & 4.7063 & \underline{4.7784} & 4.0412 & 4.1477 & \textbf{4.9123} & \textcolor{gray}{2.5227} & 4.1329 \\
     \rule{0pt}{2ex}
     & G-Eval Consist. & 4.5792	& \underline{4.7525} & 3.7823 & 3.6803 & \textbf{4.8194} & \textcolor{gray}{1.8844} & 3.8675 \\
     \rule{0pt}{2ex}
     & BARTScore+ & -2.6276 & -3.0230 & \textbf{-2.3903} & \underline{-2.5178} & -2.6570 & \textcolor{gray}{-3.3689} & -2.8863 \\
     \rule{0pt}{2ex}
     & BARTScore++ & -3.3913 & -3.7037 & \textbf{-3.1527} & \underline{-3.3382} & -3.4710 & \textcolor{gray}{-4.1312} & -3.6542 \\
     \midrule
     \multirow{6}{*}{JAMA} & ROUGE-L & \underline{0.2316} & 0.2185 & 0.2315 & 0.2185 & \textbf{0.2357} & \textcolor{gray}{0.0576} & 0.2294 \\
     \rule{0pt}{2ex}
     & BERTScore & \underline{0.6072} & 0.6018 & 0.6041 & 0.5971 & 0.6046 & \textcolor{gray}{0.4115} & \textbf{0.6087} \\
     \rule{0pt}{2ex}
     & BLEURT &  \underline{0.3395} & \textbf{0.3432} & 0.3332  & 0.3225  & 0.3302  & \textcolor{gray}{0.3098}  & 0.3324  \\
     \rule{0pt}{2ex}
     & CTC Relev. & 0.6839 & 0.6792 & \underline{0.6854} & 0.6786 & 0.6779 & \textcolor{gray}{0.6470} & \textbf{0.7012} \\
     \rule{0pt}{2ex}
     & CTC Presv. & \textbf{0.4119} & 0.4104 & 0.4107 & 0.4102 & \underline{0.4118} & \textcolor{gray}{0.3821} & 0.4113 \\
      \rule{0pt}{2ex}
      & CTC Consist. & 0.8405 & 0.8350 & 0.8450 & 0.8393 & \textcolor{gray}{0.8332} & \textbf{0.8595} & \underline{0.8593} \\
     \rule{0pt}{2ex}
    & G-Eval Relev. & 4.7426 & \underline{4.8567} & 4.1466 & 4.3031 & \textbf{4.9278} & \textcolor{gray}{1.6965} & 4.1872 \\
     \rule{0pt}{2ex}
     & G-Eval Coh. & 4.7653 & \underline{4.8304} & 4.2189 & 4.3422 & \textbf{4.9457} & \textcolor{gray}{2.1034} & 4.2418 \\
     \rule{0pt}{2ex}
     & G-Eval Consist. & 4.7068 & \underline{4.8676} & 3.9940 & 4.1278 & \textbf{4.9175} & \textcolor{gray}{1.6243} & 4.0799 \\
     \rule{0pt}{2ex}
     & BARTScore+ & -2.9695 & -3.3528 & \textbf{-2.7134} & -2.8710 & \underline{-2.8580} & \textcolor{gray}{-3.8165} & -3.2606 \\
     \rule{0pt}{2ex}
     & BARTScore++ & -3.8410 & -4.1178 & \textbf{-3.5964} & \underline{-3.7468} & -3.7918 & \textcolor{gray}{-4.5111} & -4.1018 \\
    \bottomrule
    \end{tabular}
    \caption{Automatic evaluations of the explanations generated by the seven LLMs under zero-shot \outputr prompting on Medbullets-5 and JAMA Clinical Challenge datasets. The best results for each metric are highlighted in bold, the second-best results are underlined, while the worst results are in \textcolor{gray}{gray} color.}
    \label{tab:exp_eval}
\end{table*}

\begin{table}[t]
    \centering
    \small
    \begin{tabular}{P{1.5cm}@{\hskip 0.2cm}P{1.8cm}@{\hskip 0.2cm}P{1.5cm}@{\hskip 0.2cm}P{1.5cm}@{\hskip 0.2cm}}
    \toprule
     Model & Completeness & Correctness  & Relevance \\
     \midrule
     GPT-4 & \textbf{3.35} & \textbf{4.45} & \textbf{4.61} \\
     \rule{0pt}{2ex}
     PaLM 2 & 2.67 & 4.35 & 4.53 \\
    \bottomrule
    \end{tabular}
    \caption{Human evaluations of the explanations generated by GPT-4 and PaLM 2 on Medbullets-5.}
    \label{tab:human_eval}
\end{table}
\begin{table}[t]
    \centering
    \small
    \begin{tabular}{cccc}
    \toprule
      & Human Complet. & Corr. & Relev. \\
     \midrule
     ROUGE-L & -0.28 & -0.01 & -0.05 \\
     \rule{0pt}{2ex}
     BERTScore & -0.04 & -0.06 & 0.14 \\
     \rule{0pt}{2ex}
     BLEURT & -0.14 & 0.16 & 0.04 \\
     \rule{0pt}{2ex}
     CTC Relev. & -0.27 & -0.05 & 0.09 \\
     \rule{0pt}{2ex}
     CTC Presv. & -0.06 & -0.06 & -0.00 \\
     \rule{0pt}{2ex}
     CTC Consist. & -0.37 & -0.00 &  0.03 \\
     \rule{0pt}{2ex}
     G-Eval Relev. & -0.24 & \hlc[teal!20]{0.22} & 0.11  \\
     \rule{0pt}{2ex}
     G-Eval Coh. & -0.28 & \hlc[teal!50]{0.32} &  -0.00 \\
     \rule{0pt}{2ex}
     G-Eval Consist. & -0.34 & \hlc[teal!30]{0.26} & 0.17  \\
     \rule{0pt}{2ex}
     BARTScore+ & -0.08 & 0.17 & -0.01  \\
     \rule{0pt}{2ex}
     BARTScore++ & -0.20 & 0.13 &  -0.00 \\
    \bottomrule
    \end{tabular}
    \caption{The correlation between human and automatic evaluations on the explanations generated by GPT-4 on Medbullets-5.}
    \label{tab:human_auto_corr}
\end{table}

\subsection{Chain-of-Thought (CoT) Prompting}
\label{sec:cot}
We apply zero-shot \outputry prompting to investigate whether CoT helps model reasoning on challenging medical QA. 
\autoref{tab:pred_eval} shows that CoT prompting improves accuracy of most models on MedQA and Medbullets datasets. 
However, for JAMA Clinical Challenge, CoT only improves GPT-3.5 and Meerkat but not other models.
This suggests that the challenging clinical cases in JAMA are intrinsically more difficult for models to reason about compared to board exam questions. 
This gap in performance leaves room for future work to align model reasoning with complex clinical decision-making. 
Additionally, CoT enhances Meerkat but not MedAlpaca because the former was fine-tuned with synthetic CoT data while the latter was not. 
This suggests the potential of improving models' reasoning ability by augmenting data with corresponding reasoning paths. 
We also noticed new error types introduced by CoT prompting, as discussed in the Appendix \ref{sec:cot_error}.

\subsection{Evaluating Model Explanations}
\label{sec:exp_eval}
How well do LLMs do at generating explanations as compared to human-written explanations? 
We apply \outputr to prompt the seven LLMs to produce explanations, which explicitly explain why the correct answer is preferred over the other choices (\autoref{fig:prompt} (c)).

\subsubsection{Automatic Evaluation}
\label{sec:auto_eval}
Given reference explanations in Medbullets-5 and JAMA Clinical Challenge (\S\ref{sec:datasets}), we apply several automatic metrics (\S\ref{sec:metrics}) to evaluate model-generated explanations. 
The scores are reported in \autoref{tab:exp_eval}, where the best results under each metric are highlighted in bold, the second-best results are underlined, and the worst results are in gray. 
Since the scores across different metrics are not directly comparable, we compare the ranking of explanations generated by different models under each metric. 

MedAlpaca ranks lowest on almost all metrics, but it ranks highest on CTC Consistency. 
The reason is that MedAlpaca's outputs are invalid explanations, containing text segments from the inputs, which results in artificially high consistency scores (see an example in \autoref{tab:medalpaca_expl}). 
GPT-3.5, GPT-4, Llama 3, and Meerkat are tied, ranking best or second-best in most metrics (except for BARTScore+(+)), indicating their similar abilities in generating medical explanations. 
Notably, Meerkat, fine-tuned on synthetic data generated by GPT-4, inherits its ability to explain medical questions as well as answer them. 
Moreover, Meerkat's explanations excel in CTC Relevance on both datasets, demonstrating their superiority in covering important information from reference explanations while retaining consistency with inputs. 
GPT-3.5 and GPT-4 are generally favored by ROUGE-L, BERTScore, BLEURT, and CTC Preservation, indicating that their generated explanations are more similar to reference explanations. 
All three G-Eval metrics indicate that Llama 3 and GPT-4 produce higher quality explanations than other models on both datasets.
One disagreement is that BARTScore+ and BARTScore++ score PaLM 2 as the best. 
Nevertheless, we observe that PaLM 2's explanations discuss each option but do not precisely explain why it is correct or incorrect regarding the question. 
More qualitative analyses are in Appendix \ref{sec:quali_expl}.

Overall, these automatic metrics provide consistent evaluations of model explanations, though they exhibit slight disagreements in specific rankings. 
CTC Consistency and BARTScore+(+) fall short in identifying deficient explanations, resulting in discrepancies with other metrics. 



\subsubsection{Human Evaluation}
\label{sec:human_eval}
Despite the efficiency of automatic evaluations, they struggle to assess explanations accurately due to their diversity. 
Therefore, we turn to human evaluation for a more reliable assessment of explanation quality. 
We consider three critical properties: (1) \textbf{Completeness} refers to whether the explanation sufficiently and convincingly justifies each answer choice as correct or incorrect; (2) \textbf{Correctness} means whether the information provided in the explanation is correct; (3) \textbf{Relevance} indicates whether the explanation is relevant to the question. 
We recruit crowdworkers with a Master's or Doctorate degree in Medicine/Healthcare through Prolific\footnote{https://www.prolific.com/} to evaluate 30 randomly sampled examples from Medbullets-5, along with their explanations generated by GPT-4 and PaLM 2 under \outputr prompting. 
We collect 3 annotations per instance and compute an average score for each property. 
Please see Appendix \ref{sec:app_human_eval} for more details. 

\autoref{tab:human_eval} reports human evaluation results. 
GPT-4 outperforms PaLM 2 in all three properties, which is consistent with most of the automatic evaluation results in \autoref{tab:exp_eval}. 
We compute Pearson correlation coefficients between human and automatic evaluation scores for GPT-4's explanations, as shown in \autoref{tab:human_auto_corr}. 
While G-Eval metrics demonstrate positive correlations with Human Correctness, these correlations are not strong. 
Most automatic scores show little to no correlation with human scores.
This highlights the need to develop automatic metrics that better align with human judgments. 
More analyses of human evaluation are provided in Appendix \ref{sec:app_human_eval}, where annotators identify deficiencies of model explanations, such as incorrect and irrelevant information.

\section{ Related Work}
The emergence of LLMs has significantly transformed the medical domain, particularly in the realm of medical QA tasks. 
For example, general-purpose GPT-4 \citep{openai2023gpt} has succeeded on medical board examination questions, such as the USMLE \citep{nori2023capabilities, nori2023can}. 
Additionally, many LLMs have been fine-tuned or adapted to the medical domain, deriving their domain-specific variants, such as Med-PaLM (2) \citep{singhal2022large, singhal2023towards}, MedAlpaca \citep{han2023medalpaca}, MEDITRON \citep{chen2023meditron}, and Meerkat \citep{kim2024small}. 

These models have been evaluated on medical QA benchmarks, which mostly consist of board exam questions \citep{jin2021disease, pal2022medmcqa, hendrycks2020measuring} or general medical questions \citep{abacha2019bridging, abacha2017overview, ben2019question}. 
Answering these questions, based on textbook knowledge or online resources, may be well-suited for LLMs, which learn from extensive texts \citep{dave2023chatgpt, li2023chatbots, wang2023augmenting}. 
Other datasets constructed to facilitate research on clinical QA also fail to capture the complexity of realistic clinical situations \citep{soni2022radqa, yue2021cliniqg4qa, pampari2018emrqa}. 
 
Some attempts have been made to examine LLMs on challenging clinical cases, but either the scope (restricted to a specific domain) or scale is limited \citep{strong2023chatbot, kanjee2023accuracy, shea2023use, eriksen2023use, barile2024diagnostic, jin2024hidden}. 
Another line of work creates benchmarks to evaluate LLMs in handling long clinical texts \citep{parmar2023longbox}, following clinical instructions \citep{fleming2023medalign}, or performing multiple tasks across multiple domains \citep{he2023medeval}. 
The tasks include classification and text-to-text generation (e.g., summarization, translation). 
Differently, our focus is on multiple-choice QA tasks, assessing the abilities of LLMs in applying medical reasoning to infer correct answers. 
Recent benchmarks collect questions derived from real-world conversations or closely aligned with real-world practice \citep{manes2024k, liu2024large, wang2024assessing}. 
However, these datasets do not include medical explanations. 
While MedMCQA \citep{pal2022medmcqa} provides explanations, the quality is not satisfactory. 
ExplainCPE's explanations are for Chinese Pharmacist Examination \citep{li-etal-2023-explaincpe}. 
MedExQA provides explanations for questions based on mock tests or online exams \citep{kim-etal-2024-medexqa}.
Our datasets include questions based on challenging real-world clinical cases, accompanied by high-quality, expert-written explanations. 

\section{Conclusion}
We introduce two challenging medical QA datasets, JAMA Clinical Challenge and Medbullets, accompanied by expert-written explanations. 
We evaluate seven LLMs' ability in answering and explaining these medical questions. 
We find that these datasets are harder than previous benchmarks. 
Since these are more reflective of complex clinical cases, they represent a new challenge for medical LLM research. 
We assess model-generated explanations using various automatic metrics and human evaluation. 
While LLMs produce promising explanations, they also exhibit deficiencies such as irrelevance and errors. 
Additionally, existing automatic evaluations do not correlate well with human judgements, suggesting future work to define a suite of evaluation metrics suitable for explainable medical QA. 




\section*{Limitations} 
One limitation is that we did not consider more complex prompting strategies, such as ensembling techniques \citep{wang2022self} and dynamic few-shot selection \citep{nori2023can}, which may lead to better performance on benchmarks but also add computational complexity. 
Besides, we did not apply few-shot CoT prompting using reference explanations as exemplars because they are not in the format of CoT reasoning. 
Future work might collect expert-written CoT exemplars and explore prompt engineering to enhance model performance on our constructed challenging tasks. 
Another limitation is that we excluded all figures in our collected data to tailor the task for text-based LLMs. 
These figures (e.g., X-ray imaging) might contain important information for clinical decisions. 
In the future, we will evaluate multimodal models (e.g., GPT-4V, Gemini) on the complete datasets, including both texts and images.

\section*{Ethics Statement}
The data collected in this work has been de-identified, without disclosing any sensitive information (e.g., personal identifiers). 
While model-generated explanations may contain harmful or bias information, we did not encounter such examples in our experiments. 
In human evaluation, we did not collect any personal information (e.g. demographics and identities).

\section*{Acknowledgements}
We thank Isabel Cachola for assistance in setting up the human evaluation platform and Sharon Levy for providing advice in human studies on Prolific. We thank the anonymous reviewers for many valuable comments.

\bibliography{custom}

\begin{thebibliography}{72}
\expandafter\ifx\csname natexlab\endcsname\relax\def\natexlab#1{#1}\fi

\bibitem[{Abacha et~al.(2017)Abacha, Agichtein, Pinter, and
  Demner-Fushman}]{abacha2017overview}
Asma~Ben Abacha, Eugene Agichtein, Yuval Pinter, and Dina Demner-Fushman. 2017.
\newblock \href {https://api.semanticscholar.org/CorpusID:3902472} {Overview of
  the medical question answering task at trec 2017 liveqa}.
\newblock In \emph{Text Retrieval Conference}.

\bibitem[{Abacha and Demner-Fushman(2019)}]{ben2019question}
Asma~Ben Abacha and Dina Demner-Fushman. 2019.
\newblock \href {https://api.semanticscholar.org/CorpusID:59222825} {A
  question-entailment approach to question answering}.
\newblock \emph{BMC Bioinformatics}, 20.

\bibitem[{Abacha et~al.(2019)Abacha, Mrabet, Sharp, Goodwin, Shooshan, and
  Demner-Fushman}]{abacha2019bridging}
Asma~Ben Abacha, Yassine Mrabet, Mark~E. Sharp, Travis~R. Goodwin, Sonya~E.
  Shooshan, and Dina Demner-Fushman. 2019.
\newblock \href {https://api.semanticscholar.org/CorpusID:201617660} {Bridging
  the gap between consumers' medication questions and trusted answers}.
\newblock \emph{Studies in health technology and informatics}, 264:25--29.

\bibitem[{Almazrouei et~al.(2023)Almazrouei, Alobeidli, Alshamsi, Cappelli,
  Cojocaru, Debbah, Goffinet, Hesslow, Launay, Malartic
  et~al.}]{almazrouei2023falcon}
Ebtesam Almazrouei, Hamza Alobeidli, Abdulaziz Alshamsi, Alessandro Cappelli,
  Ruxandra Cojocaru, M{\'e}rouane Debbah, {\'E}tienne Goffinet, Daniel Hesslow,
  Julien Launay, Quentin Malartic, et~al. 2023.
\newblock The falcon series of open language models.
\newblock \emph{arXiv preprint arXiv:2311.16867}.

\bibitem[{Alzahrani et~al.(2024)Alzahrani, Alyahya, Alnumay, AlRashed,
  Alsubaie, Almushayqih, Mirza, Alotaibi, Al-Twairesh, Alowisheq, Bari, and
  Khan}]{alzahrani-etal-2024-benchmarks}
Norah Alzahrani, Hisham Alyahya, Yazeed Alnumay, Sultan AlRashed, Shaykhah
  Alsubaie, Yousef Almushayqih, Faisal Mirza, Nouf Alotaibi, Nora Al-Twairesh,
  Areeb Alowisheq, M~Saiful Bari, and Haidar Khan. 2024.
\newblock \href {https://doi.org/10.18653/v1/2024.acl-long.744} {When
  benchmarks are targets: Revealing the sensitivity of large language model
  leaderboards}.
\newblock In \emph{Proceedings of the 62nd Annual Meeting of the Association
  for Computational Linguistics (Volume 1: Long Papers)}, pages 13787--13805,
  Bangkok, Thailand. Association for Computational Linguistics.

\bibitem[{Anil et~al.(2023)Anil, Dai, Firat, Johnson, Lepikhin, Passos,
  Shakeri, Taropa, Bailey, Chen et~al.}]{anil2023palm}
Rohan Anil, Andrew~M Dai, Orhan Firat, Melvin Johnson, Dmitry Lepikhin,
  Alexandre Passos, Siamak Shakeri, Emanuel Taropa, Paige Bailey, Zhifeng Chen,
  et~al. 2023.
\newblock Palm 2 technical report.
\newblock \emph{arXiv preprint arXiv:2305.10403}.

\bibitem[{Anthropic(2024)}]{anthropicclaude3}
Anthropic. 2024.
\newblock \href {https://api.semanticscholar.org/CorpusID:268232499} {The
  claude 3 model family: Opus, sonnet, haiku}.

\bibitem[{Barile et~al.(2024)Barile, Margolis, Cason, Kim, Kalash, Tchaconas,
  and Milanaik}]{barile2024diagnostic}
Joseph Barile, Alex Margolis, Grace Cason, Rachel Kim, Saia Kalash, Alexis
  Tchaconas, and Ruth Milanaik. 2024.
\newblock \href {https://doi.org/10.1001/jamapediatrics.2023.5750} {Diagnostic
  accuracy of a large language model in pediatric case studies}.
\newblock \emph{JAMA Pediatrics}, 178(3):313--315.

\bibitem[{BENNETT et~al.(1954)BENNETT, ALPERT, and
  GOLDSTEIN}]{bennett1954communications}
E.~M. BENNETT, R.~ALPERT, and A.~C. GOLDSTEIN. 1954.
\newblock \href {https://doi.org/10.1086/266520} {Communications through
  limited-response questioning}.
\newblock \emph{Public Opinion Quarterly}, 18(3):303--308.

\bibitem[{Brown et~al.(2020)Brown, Mann, Ryder, Subbiah, Kaplan, Dhariwal,
  Neelakantan, Shyam, Sastry, Askell, Agarwal, Herbert-Voss, Krueger, Henighan,
  Child, Ramesh, Ziegler, Wu, Winter, Hesse, Chen, Sigler, Litwin, Gray, Chess,
  Clark, Berner, McCandlish, Radford, Sutskever, and
  Amodei}]{brown2020language}
Tom Brown, Benjamin Mann, Nick Ryder, Melanie Subbiah, Jared~D Kaplan, Prafulla
  Dhariwal, Arvind Neelakantan, Pranav Shyam, Girish Sastry, Amanda Askell,
  Sandhini Agarwal, Ariel Herbert-Voss, Gretchen Krueger, Tom Henighan, Rewon
  Child, Aditya Ramesh, Daniel Ziegler, Jeffrey Wu, Clemens Winter, Chris
  Hesse, Mark Chen, Eric Sigler, Mateusz Litwin, Scott Gray, Benjamin Chess,
  Jack Clark, Christopher Berner, Sam McCandlish, Alec Radford, Ilya Sutskever,
  and Dario Amodei. 2020.
\newblock \href
  {https://proceedings.neurips.cc/paper_files/paper/2020/file/1457c0d6bfcb4967418bfb8ac142f64a-Paper.pdf}
  {Language models are few-shot learners}.
\newblock In \emph{Advances in Neural Information Processing Systems},
  volume~33, pages 1877--1901. Curran Associates, Inc.

\bibitem[{Chen et~al.(2022)Chen, Zhong, Zha, Karypis, and
  He}]{chen-etal-2022-meta}
Yanda Chen, Ruiqi Zhong, Sheng Zha, George Karypis, and He~He. 2022.
\newblock \href {https://doi.org/10.18653/v1/2022.acl-long.53} {Meta-learning
  via language model in-context tuning}.
\newblock In \emph{Proceedings of the 60th Annual Meeting of the Association
  for Computational Linguistics (Volume 1: Long Papers)}, pages 719--730,
  Dublin, Ireland. Association for Computational Linguistics.

\bibitem[{Chen et~al.(2023)Chen, Cano, Romanou, Bonnet, Matoba, Salvi,
  Pagliardini, Fan, Kopf, Mohtashami, Sallinen, Sakhaeirad, Swamy, Krawczuk,
  Bayazit, Marmet, Montariol, Hartley, Jaggi, and Bosselut}]{chen2023meditron}
Zeming Chen, Alejandro~Hern'andez Cano, Angelika Romanou, Antoine Bonnet, Kyle
  Matoba, Francesco Salvi, Matteo Pagliardini, Simin Fan, Andreas Kopf,
  Amirkeivan Mohtashami, Alexandre Sallinen, Alireza Sakhaeirad, Vinitra Swamy,
  Igor Krawczuk, Deniz Bayazit, Axel Marmet, Syrielle Montariol, Mary-Anne
  Hartley, Martin Jaggi, and Antoine Bosselut. 2023.
\newblock \href {https://api.semanticscholar.org/CorpusID:265456229}
  {Meditron-70b: Scaling medical pretraining for large language models}.
\newblock \emph{ArXiv}, abs/2311.16079.

\bibitem[{Chowdhery et~al.(2023)Chowdhery, Narang, Devlin, Bosma, Mishra,
  Roberts, Barham, Chung, Sutton, Gehrmann, Schuh, Shi, Tsvyashchenko, Maynez,
  Rao, Barnes, Tay, Shazeer, Prabhakaran, Reif, Du, Hutchinson, Pope, Bradbury,
  Austin, Isard, Gur-Ari, Yin, Duke, Levskaya, Ghemawat, Dev, Michalewski,
  Garcia, Misra, Robinson, Fedus, Zhou, Ippolito, Luan, Lim, Zoph, Spiridonov,
  Sepassi, Dohan, Agrawal, Omernick, Dai, Pillai, Pellat, Lewkowycz, Moreira,
  Child, Polozov, Lee, Zhou, Wang, Saeta, Diaz, Firat, Catasta, Wei,
  Meier-Hellstern, Eck, Dean, Petrov, and Fiedel}]{chowdhery2023palm}
Aakanksha Chowdhery, Sharan Narang, Jacob Devlin, Maarten Bosma, Gaurav Mishra,
  Adam Roberts, Paul Barham, Hyung~Won Chung, Charles Sutton, Sebastian
  Gehrmann, Parker Schuh, Kensen Shi, Sashank Tsvyashchenko, Joshua Maynez,
  Abhishek Rao, Parker Barnes, Yi~Tay, Noam Shazeer, Vinodkumar Prabhakaran,
  Emily Reif, Nan Du, Ben Hutchinson, Reiner Pope, James Bradbury, Jacob
  Austin, Michael Isard, Guy Gur-Ari, Pengcheng Yin, Toju Duke, Anselm
  Levskaya, Sanjay Ghemawat, Sunipa Dev, Henryk Michalewski, Xavier Garcia,
  Vedant Misra, Kevin Robinson, Liam Fedus, Denny Zhou, Daphne Ippolito, David
  Luan, Hyeontaek Lim, Barret Zoph, Alexander Spiridonov, Ryan Sepassi, David
  Dohan, Shivani Agrawal, Mark Omernick, Andrew~M. Dai,
  Thanumalayan~Sankaranarayana Pillai, Marie Pellat, Aitor Lewkowycz, Erica
  Moreira, Rewon Child, Oleksandr Polozov, Katherine Lee, Zongwei Zhou, Xuezhi
  Wang, Brennan Saeta, Mark Diaz, Orhan Firat, Michele Catasta, Jason Wei,
  Kathy Meier-Hellstern, Douglas Eck, Jeff Dean, Slav Petrov, and Noah Fiedel.
  2023.
\newblock Palm: scaling language modeling with pathways.
\newblock \emph{J. Mach. Learn. Res.}, 24(1).

\bibitem[{Dave et~al.(2023)Dave, Athaluri, and Singh}]{dave2023chatgpt}
Tirth Dave, Sai~Anirudh Athaluri, and Satyam Singh. 2023.
\newblock Chatgpt in medicine: an overview of its applications, advantages,
  limitations, future prospects, and ethical considerations.
\newblock \emph{Frontiers in Artificial Intelligence}, 6:1169595.

\bibitem[{Deng et~al.(2024)Deng, Zhao, Tang, Gerstein, and
  Cohan}]{deng-etal-2024-investigating}
Chunyuan Deng, Yilun Zhao, Xiangru Tang, Mark Gerstein, and Arman Cohan. 2024.
\newblock \href {https://doi.org/10.18653/v1/2024.naacl-long.482}
  {Investigating data contamination in modern benchmarks for large language
  models}.
\newblock In \emph{Proceedings of the 2024 Conference of the North American
  Chapter of the Association for Computational Linguistics: Human Language
  Technologies (Volume 1: Long Papers)}, pages 8706--8719, Mexico City, Mexico.
  Association for Computational Linguistics.

\bibitem[{Deng et~al.(2021)Deng, Tan, Liu, Xing, and
  Hu}]{deng-etal-2021-compression}
Mingkai Deng, Bowen Tan, Zhengzhong Liu, Eric Xing, and Zhiting Hu. 2021.
\newblock \href {https://doi.org/10.18653/v1/2021.emnlp-main.599} {Compression,
  transduction, and creation: A unified framework for evaluating natural
  language generation}.
\newblock In \emph{Proceedings of the 2021 Conference on Empirical Methods in
  Natural Language Processing}, pages 7580--7605, Online and Punta Cana,
  Dominican Republic. Association for Computational Linguistics.

\bibitem[{Devlin et~al.(2019)Devlin, Chang, Lee, and
  Toutanova}]{devlin-etal-2019-bert}
Jacob Devlin, Ming-Wei Chang, Kenton Lee, and Kristina Toutanova. 2019.
\newblock \href {https://doi.org/10.18653/v1/N19-1423} {{BERT}: Pre-training of
  deep bidirectional transformers for language understanding}.
\newblock In \emph{Proceedings of the 2019 Conference of the North {A}merican
  Chapter of the Association for Computational Linguistics: Human Language
  Technologies, Volume 1 (Long and Short Papers)}, pages 4171--4186,
  Minneapolis, Minnesota. Association for Computational Linguistics.

\bibitem[{Eriksen et~al.(2024)Eriksen, Möller, and Ryg}]{eriksen2023use}
Alexander~V. Eriksen, Sören Möller, and Jesper Ryg. 2024.
\newblock \href {https://doi.org/10.1056/AIp2300031} {Use of gpt-4 to diagnose
  complex clinical cases}.
\newblock \emph{NEJM AI}, 1(1):AIp2300031.

\bibitem[{Fleming et~al.(2024)Fleming, Lozano, Haberkorn, Jindal, Reis, Thapa,
  Blankemeier, Genkins, Steinberg, Nayak, Patel, Chiang, Callahan, Huo,
  Gatidis, Adams, Fayanju, Shah, Savage, Goh, Chaudhari, Aghaeepour, Sharp,
  Pfeffer, Liang, Chen, Morse, Brunskill, Fries, and
  Shah}]{fleming2023medalign}
Scott~L. Fleming, Alejandro Lozano, William~J. Haberkorn, Jenelle~A. Jindal,
  Eduardo~Pontes Reis, Rahul Thapa, Louis Blankemeier, Julian~Z. Genkins, Ethan
  Steinberg, Ashwin Nayak, Birju~S. Patel, Chia{-}Chun Chiang, Alison Callahan,
  Zepeng Huo, Sergios Gatidis, Scott~J. Adams, Oluseyi Fayanju, Shreya~J. Shah,
  Thomas Savage, Ethan Goh, Akshay~S. Chaudhari, Nima Aghaeepour,
  Christopher~D. Sharp, Michael~A. Pfeffer, Percy Liang, Jonathan~H. Chen,
  Keith~E. Morse, Emma~P. Brunskill, Jason~A. Fries, and Nigam~H. Shah. 2024.
\newblock \href {https://doi.org/10.1609/AAAI.V38I20.30205} {Medalign: {A}
  clinician-generated dataset for instruction following with electronic medical
  records}.
\newblock In \emph{Thirty-Eighth {AAAI} Conference on Artificial Intelligence,
  {AAAI} 2024, Thirty-Sixth Conference on Innovative Applications of Artificial
  Intelligence, {IAAI} 2024, Fourteenth Symposium on Educational Advances in
  Artificial Intelligence, {EAAI} 2014, February 20-27, 2024, Vancouver,
  Canada}, pages 22021--22030. {AAAI} Press.

\bibitem[{Han et~al.(2023)Han, Adams, Papaioannou, Grundmann, Oberhauser,
  L{\"o}ser, Truhn, and Bressem}]{han2023medalpaca}
Tianyu Han, Lisa~C Adams, Jens-Michalis Papaioannou, Paul Grundmann, Tom
  Oberhauser, Alexander L{\"o}ser, Daniel Truhn, and Keno~K Bressem. 2023.
\newblock Medalpaca--an open-source collection of medical conversational ai
  models and training data.
\newblock \emph{arXiv preprint arXiv:2304.08247}.

\bibitem[{Harris(2023)}]{harris2023large}
Emily Harris. 2023.
\newblock \href {https://doi.org/10.1001/jama.2023.14311} {Large language
  models answer medical questions accurately, but can’t match clinicians’
  knowledge}.
\newblock \emph{JAMA}, 330(9):792--794.

\bibitem[{Hastie et~al.(2009)Hastie, Tibshirani, Friedman, and
  Friedman}]{hastie2009elements}
Trevor Hastie, Robert Tibshirani, Jerome~H Friedman, and Jerome~H Friedman.
  2009.
\newblock \emph{The elements of statistical learning: data mining, inference,
  and prediction}, volume~2.
\newblock Springer.

\bibitem[{He et~al.(2023)He, Wang, Yan, Liu, Chang, Gentili, McAuley, and
  Hsu}]{he2023medeval}
Zexue He, Yu~Wang, An~Yan, Yao Liu, Eric Chang, Amilcare Gentili, Julian
  McAuley, and Chun-Nan Hsu. 2023.
\newblock \href {https://doi.org/10.18653/v1/2023.emnlp-main.540} {{M}ed{E}val:
  A multi-level, multi-task, and multi-domain medical benchmark for language
  model evaluation}.
\newblock In \emph{Proceedings of the 2023 Conference on Empirical Methods in
  Natural Language Processing}, pages 8725--8744, Singapore. Association for
  Computational Linguistics.

\bibitem[{Hendrycks et~al.(2020)Hendrycks, Burns, Basart, Zou, Mazeika, Song,
  and Steinhardt}]{hendrycks2020measuring}
Dan Hendrycks, Collin Burns, Steven Basart, Andy Zou, Mantas Mazeika, Dawn
  Song, and Jacob Steinhardt. 2020.
\newblock Measuring massive multitask language understanding.
\newblock \emph{arXiv preprint arXiv:2009.03300}.

\bibitem[{Hermann et~al.(2015)Hermann, Kocisky, Grefenstette, Espeholt, Kay,
  Suleyman, and Blunsom}]{hermann2015teaching}
Karl~Moritz Hermann, Tomas Kocisky, Edward Grefenstette, Lasse Espeholt, Will
  Kay, Mustafa Suleyman, and Phil Blunsom. 2015.
\newblock Teaching machines to read and comprehend.
\newblock \emph{Advances in neural information processing systems}, 28.

\bibitem[{Hu et~al.(2019)Hu, Singh, Holzenberger, Post, and
  Van~Durme}]{hu2019large}
J.~Edward Hu, Abhinav Singh, Nils Holzenberger, Matt Post, and Benjamin
  Van~Durme. 2019.
\newblock \href {https://doi.org/10.18653/v1/K19-1005} {Large-scale, diverse,
  paraphrastic bitexts via sampling and clustering}.
\newblock In \emph{Proceedings of the 23rd Conference on Computational Natural
  Language Learning (CoNLL)}, pages 44--54, Hong Kong, China. Association for
  Computational Linguistics.

\bibitem[{Jin et~al.(2021)Jin, Pan, Oufattole, Weng, Fang, and
  Szolovits}]{jin2021disease}
Di~Jin, Eileen Pan, Nassim Oufattole, Wei-Hung Weng, Hanyi Fang, and Peter
  Szolovits. 2021.
\newblock What disease does this patient have? a large-scale open domain
  question answering dataset from medical exams.
\newblock \emph{Applied Sciences}, 11(14):6421.

\bibitem[{Jin et~al.(2024)Jin, Chen, Zhou, Xu, Cheung, Chen, Summers, Rousseau,
  Ni, Landsman et~al.}]{jin2024hidden}
Qiao Jin, Fangyuan Chen, Yiliang Zhou, Ziyang Xu, Justin~M Cheung, Robert Chen,
  Ronald~M Summers, Justin~F Rousseau, Peiyun Ni, Marc~J Landsman, et~al. 2024.
\newblock Hidden flaws behind expert-level accuracy of gpt-4 vision in
  medicine.
\newblock \emph{arXiv preprint arXiv:2401.08396}.

\bibitem[{Kanjee et~al.(2023)Kanjee, Crowe, and Rodman}]{kanjee2023accuracy}
Zahir Kanjee, Byron Crowe, and Adam Rodman. 2023.
\newblock Accuracy of a generative artificial intelligence model in a complex
  diagnostic challenge.
\newblock \emph{JAMA}.

\bibitem[{Kim et~al.(2024{\natexlab{a}})Kim, Hwang, Lee, Park, Kim, Lee, Yoon,
  Sohn, Choi, and Kang}]{kim2024small}
Hyunjae Kim, Hyeon Hwang, Jiwoo Lee, Sihyeon Park, Dain Kim, Taewhoo Lee,
  Chanwoong Yoon, Jiwoong Sohn, Donghee Choi, and Jaewoo Kang.
  2024{\natexlab{a}}.
\newblock Small language models learn enhanced reasoning skills from medical
  textbooks.
\newblock \emph{arXiv preprint arXiv:2404.00376}.

\bibitem[{Kim et~al.(2024{\natexlab{b}})Kim, Wu, Abdulle, and
  Wu}]{kim-etal-2024-medexqa}
Yunsoo Kim, Jinge Wu, Yusuf Abdulle, and Honghan Wu. 2024{\natexlab{b}}.
\newblock \href {https://doi.org/10.18653/v1/2024.bionlp-1.14} {{M}ed{E}x{QA}:
  Medical question answering benchmark with multiple explanations}.
\newblock In \emph{Proceedings of the 23rd Workshop on Biomedical Natural
  Language Processing}, pages 167--181, Bangkok, Thailand. Association for
  Computational Linguistics.

\bibitem[{Kojima et~al.(2022)Kojima, Gu, Reid, Matsuo, and
  Iwasawa}]{kojima2022large}
Takeshi Kojima, Shixiang~Shane Gu, Machel Reid, Yutaka Matsuo, and Yusuke
  Iwasawa. 2022.
\newblock Large language models are zero-shot reasoners.
\newblock \emph{Advances in neural information processing systems},
  35:22199--22213.

\bibitem[{Kweon et~al.(2024)Kweon, Kim, Kim, Im, Cho, Bae, Oh, Lee, Moon, You,
  Baek, Han, Jung, Jo, and Choi}]{kweon-etal-2024-publicly}
Sunjun Kweon, Junu Kim, Jiyoun Kim, Sujeong Im, Eunbyeol Cho, Seongsu Bae,
  Jungwoo Oh, Gyubok Lee, Jong~Hak Moon, Seng~Chan You, Seungjin Baek,
  Chang~Hoon Han, Yoon~Bin Jung, Yohan Jo, and Edward Choi. 2024.
\newblock \href {https://doi.org/10.18653/v1/2024.findings-acl.305} {Publicly
  shareable clinical large language model built on synthetic clinical notes}.
\newblock In \emph{Findings of the Association for Computational Linguistics
  ACL 2024}, pages 5148--5168, Bangkok, Thailand and virtual meeting.
  Association for Computational Linguistics.

\bibitem[{Lewis et~al.(2020)Lewis, Liu, Goyal, Ghazvininejad, Mohamed, Levy,
  Stoyanov, and Zettlemoyer}]{lewis-etal-2020-bart}
Mike Lewis, Yinhan Liu, Naman Goyal, Marjan Ghazvininejad, Abdelrahman Mohamed,
  Omer Levy, Veselin Stoyanov, and Luke Zettlemoyer. 2020.
\newblock \href {https://doi.org/10.18653/v1/2020.acl-main.703} {{BART}:
  Denoising sequence-to-sequence pre-training for natural language generation,
  translation, and comprehension}.
\newblock In \emph{Proceedings of the 58th Annual Meeting of the Association
  for Computational Linguistics}, pages 7871--7880, Online. Association for
  Computational Linguistics.

\bibitem[{Li et~al.(2023{\natexlab{a}})Li, Yu, Hu, Xu, and
  Zhang}]{li-etal-2023-explaincpe}
Dongfang Li, Jindi Yu, Baotian Hu, Zhenran Xu, and Min Zhang.
  2023{\natexlab{a}}.
\newblock \href {https://doi.org/10.18653/v1/2023.findings-emnlp.129}
  {{E}xplain{CPE}: A free-text explanation benchmark of {C}hinese pharmacist
  examination}.
\newblock In \emph{Findings of the Association for Computational Linguistics:
  EMNLP 2023}, pages 1922--1940, Singapore. Association for Computational
  Linguistics.

\bibitem[{Li et~al.(2023{\natexlab{b}})Li, Kumar, and Chen}]{li2023chatbots}
Ron Li, Andre Kumar, and Jonathan~H Chen. 2023{\natexlab{b}}.
\newblock How chatbots and large language model artificial intelligence systems
  will reshape modern medicine: Fountain of creativity or pandora’s box?
\newblock \emph{JAMA Internal Medicine}.

\bibitem[{Li{\'e}vin et~al.(2022)Li{\'e}vin, Hother, and
  Winther}]{lievin2022can}
Valentin Li{\'e}vin, Christoffer~Egeberg Hother, and Ole Winther. 2022.
\newblock Can large language models reason about medical questions?
\newblock \emph{arXiv preprint arXiv:2207.08143}.

\bibitem[{Lin(2004)}]{lin2004rouge}
Chin-Yew Lin. 2004.
\newblock Rouge: A package for automatic evaluation of summaries.
\newblock In \emph{Text summarization branches out}, pages 74--81.

\bibitem[{Liu et~al.(2024)Liu, Li, Zhou, Yin, Yang, Tang, Luo, Zeng, Jiang,
  Gao, Nigam, Nag, Yin, Hua, Zhou, Rohanian, Thakur, Clifton, and
  Clifton}]{liu2024large}
Fenglin Liu, Zheng Li, Hongjian Zhou, Qingyu Yin, Jingfeng Yang, Xianfeng Tang,
  Chen Luo, Ming Zeng, Haoming Jiang, Yifan Gao, Priyanka Nigam, Sreyashi Nag,
  Bing Yin, Yining Hua, Xuan Zhou, Omid Rohanian, Anshul Thakur, Lei Clifton,
  and David~A. Clifton. 2024.
\newblock \href {https://doi.org/10.18653/v1/2024.emnlp-main.759} {Large
  language models are poor clinical decision-makers: A comprehensive
  benchmark}.
\newblock In \emph{Proceedings of the 2024 Conference on Empirical Methods in
  Natural Language Processing}, pages 13696--13710, Miami, Florida, USA.
  Association for Computational Linguistics.

\bibitem[{Liu et~al.(2023)Liu, Iter, Xu, Wang, Xu, and Zhu}]{liu-etal-2023-g}
Yang Liu, Dan Iter, Yichong Xu, Shuohang Wang, Ruochen Xu, and Chenguang Zhu.
  2023.
\newblock \href {https://doi.org/10.18653/v1/2023.emnlp-main.153} {{G}-eval:
  {NLG} evaluation using gpt-4 with better human alignment}.
\newblock In \emph{Proceedings of the 2023 Conference on Empirical Methods in
  Natural Language Processing}, pages 2511--2522, Singapore. Association for
  Computational Linguistics.

\bibitem[{Liu et~al.(2019)Liu, Ott, Goyal, Du, Joshi, Chen, Levy, Lewis,
  Zettlemoyer, and Stoyanov}]{liu2019roberta}
Yinhan Liu, Myle Ott, Naman Goyal, Jingfei Du, Mandar Joshi, Danqi Chen, Omer
  Levy, Mike Lewis, Luke Zettlemoyer, and Veselin Stoyanov. 2019.
\newblock Roberta: A robustly optimized bert pretraining approach.
\newblock \emph{arXiv preprint arXiv:1907.11692}.

\bibitem[{Magar and Schwartz(2022)}]{magar-schwartz-2022-data}
Inbal Magar and Roy Schwartz. 2022.
\newblock \href {https://doi.org/10.18653/v1/2022.acl-short.18} {Data
  contamination: From memorization to exploitation}.
\newblock In \emph{Proceedings of the 60th Annual Meeting of the Association
  for Computational Linguistics (Volume 2: Short Papers)}, pages 157--165,
  Dublin, Ireland. Association for Computational Linguistics.

\bibitem[{Manes et~al.(2024)Manes, Ronn, Cohen, Ber, Horowitz-Kugler, and
  Stanovsky}]{manes2024k}
Itay Manes, Naama Ronn, David Cohen, Ran~Ilan Ber, Zehavi Horowitz-Kugler, and
  Gabriel Stanovsky. 2024.
\newblock K-qa: A real-world medical q\&a benchmark.
\newblock \emph{arXiv preprint arXiv:2401.14493}.

\bibitem[{Meta(2024)}]{meta2024llama}
Meta. 2024.
\newblock Introducing meta llama 3: The most capable openly available llm to
  date.

\bibitem[{Min et~al.(2022)Min, Lyu, Holtzman, Artetxe, Lewis, Hajishirzi, and
  Zettlemoyer}]{min-etal-2022-rethinking}
Sewon Min, Xinxi Lyu, Ari Holtzman, Mikel Artetxe, Mike Lewis, Hannaneh
  Hajishirzi, and Luke Zettlemoyer. 2022.
\newblock \href {https://doi.org/10.18653/v1/2022.emnlp-main.759} {Rethinking
  the role of demonstrations: What makes in-context learning work?}
\newblock In \emph{Proceedings of the 2022 Conference on Empirical Methods in
  Natural Language Processing}, pages 11048--11064, Abu Dhabi, United Arab
  Emirates. Association for Computational Linguistics.

\bibitem[{Nori et~al.(2023{\natexlab{a}})Nori, King, McKinney, Carignan, and
  Horvitz}]{nori2023capabilities}
Harsha Nori, Nicholas King, Scott~Mayer McKinney, Dean Carignan, and Eric
  Horvitz. 2023{\natexlab{a}}.
\newblock Capabilities of gpt-4 on medical challenge problems.
\newblock \emph{arXiv preprint arXiv:2303.13375}.

\bibitem[{Nori et~al.(2023{\natexlab{b}})Nori, Lee, Zhang, Carignan, Edgar,
  Fusi, King, Larson, Li, Liu et~al.}]{nori2023can}
Harsha Nori, Yin~Tat Lee, Sheng Zhang, Dean Carignan, Richard Edgar, Nicolo
  Fusi, Nicholas King, Jonathan Larson, Yuanzhi Li, Weishung Liu, et~al.
  2023{\natexlab{b}}.
\newblock Can generalist foundation models outcompete special-purpose tuning?
  case study in medicine.
\newblock \emph{arXiv preprint arXiv:2311.16452}.

\bibitem[{OpenAI(2023)}]{openai2023gpt}
OpenAI. 2023.
\newblock Gpt-4 technical report.

\bibitem[{Ouyang et~al.(2022)Ouyang, Wu, Jiang, Almeida, Wainwright, Mishkin,
  Zhang, Agarwal, Slama, Ray, Schulman, Hilton, Kelton, Miller, Simens, Askell,
  Welinder, Christiano, Leike, and Lowe}]{ouyang2022training}
Long Ouyang, Jeff Wu, Xu~Jiang, Diogo Almeida, Carroll~L. Wainwright, Pamela
  Mishkin, Chong Zhang, Sandhini Agarwal, Katarina Slama, Alex Ray, John
  Schulman, Jacob Hilton, Fraser Kelton, Luke Miller, Maddie Simens, Amanda
  Askell, Peter Welinder, Paul Christiano, Jan Leike, and Ryan Lowe. 2022.
\newblock Training language models to follow instructions with human feedback.
\newblock In \emph{Proceedings of the 36th International Conference on Neural
  Information Processing Systems}, NIPS '22, Red Hook, NY, USA. Curran
  Associates Inc.

\bibitem[{Pal et~al.(2022)Pal, Umapathi, and Sankarasubbu}]{pal2022medmcqa}
Ankit Pal, Logesh~Kumar Umapathi, and Malaikannan Sankarasubbu. 2022.
\newblock \href {https://proceedings.mlr.press/v174/pal22a.html} {Medmcqa: A
  large-scale multi-subject multi-choice dataset for medical domain question
  answering}.
\newblock In \emph{Proceedings of the Conference on Health, Inference, and
  Learning}, volume 174 of \emph{Proceedings of Machine Learning Research},
  pages 248--260. PMLR.

\bibitem[{Pampari et~al.(2018)Pampari, Raghavan, Liang, and
  Peng}]{pampari2018emrqa}
Anusri Pampari, Preethi Raghavan, Jennifer Liang, and Jian Peng. 2018.
\newblock emrqa: A large corpus for question answering on electronic medical
  records.
\newblock \emph{arXiv preprint arXiv:1809.00732}.

\bibitem[{Panigutti et~al.(2022)Panigutti, Beretta, Giannotti, and
  Pedreschi}]{panigutti2022understanding}
Cecilia Panigutti, Andrea Beretta, Fosca Giannotti, and Dino Pedreschi. 2022.
\newblock \href {https://doi.org/10.1145/3491102.3502104} {Understanding the
  impact of explanations on advice-taking: a user study for ai-based clinical
  decision support systems}.
\newblock In \emph{Proceedings of the 2022 CHI Conference on Human Factors in
  Computing Systems}, CHI '22, New York, NY, USA. Association for Computing
  Machinery.

\bibitem[{Parmar et~al.(2023)Parmar, Naik, Gupta, Agrawal, and
  Baral}]{parmar2023longbox}
Mihir Parmar, Aakanksha Naik, Himanshu Gupta, Disha Agrawal, and Chitta Baral.
  2023.
\newblock Longbox: Evaluating transformers on long-sequence clinical tasks.
\newblock \emph{arXiv preprint arXiv:2311.09564}.

\bibitem[{Saab et~al.(2024)Saab, Tu, Weng, Tanno, Stutz, Wulczyn, Zhang,
  Strother, Park, Vedadi et~al.}]{saab2024capabilities}
Khaled Saab, Tao Tu, Wei-Hung Weng, Ryutaro Tanno, David Stutz, Ellery Wulczyn,
  Fan Zhang, Tim Strother, Chunjong Park, Elahe Vedadi, et~al. 2024.
\newblock Capabilities of gemini models in medicine.
\newblock \emph{arXiv preprint arXiv:2404.18416}.

\bibitem[{Sainz et~al.(2023)Sainz, Campos, Garc{\'\i}a-Ferrero, Etxaniz,
  de~Lacalle, and Agirre}]{sainz-etal-2023-nlp}
Oscar Sainz, Jon Campos, Iker Garc{\'\i}a-Ferrero, Julen Etxaniz, Oier~Lopez
  de~Lacalle, and Eneko Agirre. 2023.
\newblock \href {https://doi.org/10.18653/v1/2023.findings-emnlp.722} {{NLP}
  evaluation in trouble: On the need to measure {LLM} data contamination for
  each benchmark}.
\newblock In \emph{Findings of the Association for Computational Linguistics:
  EMNLP 2023}, pages 10776--10787, Singapore. Association for Computational
  Linguistics.

\bibitem[{Sellam et~al.(2020)Sellam, Das, and Parikh}]{sellam-etal-2020-bleurt}
Thibault Sellam, Dipanjan Das, and Ankur Parikh. 2020.
\newblock \href {https://doi.org/10.18653/v1/2020.acl-main.704} {{BLEURT}:
  Learning robust metrics for text generation}.
\newblock In \emph{Proceedings of the 58th Annual Meeting of the Association
  for Computational Linguistics}, pages 7881--7892, Online. Association for
  Computational Linguistics.

\bibitem[{Shea et~al.(2023)Shea, Lee, Ip, Luk, and Wong}]{shea2023use}
Yat-Fung Shea, Cynthia Min~Yao Lee, Whitney Chin~Tung Ip, Dik Wai~Anderson Luk,
  and Stephanie Sze~Wing Wong. 2023.
\newblock \href {https://doi.org/10.1001/jamanetworkopen.2023.25000} {Use of
  gpt-4 to analyze medical records of patients with extensive investigations
  and delayed diagnosis}.
\newblock \emph{JAMA Network Open}, 6(8):e2325000--e2325000.

\bibitem[{Singhal et~al.(2022)Singhal, Azizi, Tu, Mahdavi, Wei, Chung, Scales,
  Tanwani, Cole-Lewis, Pfohl et~al.}]{singhal2022large}
Karan Singhal, Shekoofeh Azizi, Tao Tu, S~Sara Mahdavi, Jason Wei, Hyung~Won
  Chung, Nathan Scales, Ajay Tanwani, Heather Cole-Lewis, Stephen Pfohl, et~al.
  2022.
\newblock Large language models encode clinical knowledge.
\newblock \emph{arXiv preprint arXiv:2212.13138}.

\bibitem[{Singhal et~al.(2023)Singhal, Tu, Gottweis, Sayres, Wulczyn, Hou,
  Clark, Pfohl, Cole-Lewis, Neal et~al.}]{singhal2023towards}
Karan Singhal, Tao Tu, Juraj Gottweis, Rory Sayres, Ellery Wulczyn, Le~Hou,
  Kevin Clark, Stephen Pfohl, Heather Cole-Lewis, Darlene Neal, et~al. 2023.
\newblock Towards expert-level medical question answering with large language
  models.
\newblock \emph{arXiv preprint arXiv:2305.09617}.

\bibitem[{Soni et~al.(2022)Soni, Gudala, Pajouhi, and Roberts}]{soni2022radqa}
Sarvesh Soni, Meghana Gudala, Atieh Pajouhi, and Kirk Roberts. 2022.
\newblock \href {https://aclanthology.org/2022.lrec-1.672/} {{R}ad{QA}: A
  question answering dataset to improve comprehension of radiology reports}.
\newblock In \emph{Proceedings of the Thirteenth Language Resources and
  Evaluation Conference}, pages 6250--6259, Marseille, France. European
  Language Resources Association.

\bibitem[{Strong et~al.(2023)Strong, DiGiammarino, Weng, Kumar, Hosamani, Hom,
  and Chen}]{strong2023chatbot}
Eric Strong, Alicia DiGiammarino, Yingjie Weng, Andre Kumar, Poonam Hosamani,
  Jason Hom, and Jonathan~H. Chen. 2023.
\newblock \href {https://doi.org/10.1001/jamainternmed.2023.2909} {Chatbot vs
  medical student performance on free-response clinical reasoning
  examinations}.
\newblock \emph{JAMA Internal Medicine}, 183(9):1028--1030.

\bibitem[{Team et~al.(2023)Team, Anil, Borgeaud, Wu, Alayrac, Yu, Soricut,
  Schalkwyk, Dai, Hauth et~al.}]{team2023gemini}
Gemini Team, Rohan Anil, Sebastian Borgeaud, Yonghui Wu, Jean-Baptiste Alayrac,
  Jiahui Yu, Radu Soricut, Johan Schalkwyk, Andrew~M Dai, Anja Hauth, et~al.
  2023.
\newblock Gemini: a family of highly capable multimodal models.
\newblock \emph{arXiv preprint arXiv:2312.11805}.

\bibitem[{Touvron et~al.(2023)Touvron, Martin, Stone, Albert, Almahairi,
  Babaei, Bashlykov, Batra, Bhargava, Bhosale et~al.}]{touvron2023llama}
Hugo Touvron, Louis Martin, Kevin Stone, Peter Albert, Amjad Almahairi, Yasmine
  Babaei, Nikolay Bashlykov, Soumya Batra, Prajjwal Bhargava, Shruti Bhosale,
  et~al. 2023.
\newblock Llama 2: Open foundation and fine-tuned chat models.
\newblock \emph{arXiv preprint arXiv:2307.09288}.

\bibitem[{Wang et~al.(2024)Wang, Ran, Tang, Chang, Chuang, Liu, Braverman, Liu,
  and Hu}]{wang2024assessing}
Guanchu Wang, Junhao Ran, Ruixiang Tang, Chia-Yuan Chang, Yu-Neng Chuang, Zirui
  Liu, Vladimir Braverman, Zhandong Liu, and Xia Hu. 2024.
\newblock Assessing and enhancing large language models in rare disease
  question-answering.
\newblock \emph{arXiv preprint arXiv:2408.08422}.

\bibitem[{Wang et~al.(2022)Wang, Wei, Schuurmans, Le, Chi, Narang, Chowdhery,
  and Zhou}]{wang2022self}
Xuezhi Wang, Jason Wei, Dale Schuurmans, Quoc Le, Ed~Chi, Sharan Narang,
  Aakanksha Chowdhery, and Denny Zhou. 2022.
\newblock Self-consistency improves chain of thought reasoning in language
  models.
\newblock \emph{arXiv preprint arXiv:2203.11171}.

\bibitem[{Wang et~al.(2023)Wang, Ma, and Chen}]{wang2023augmenting}
Yubo Wang, Xueguang Ma, and Wenhu Chen. 2023.
\newblock Augmenting black-box llms with medical textbooks for clinical
  question answering.
\newblock \emph{arXiv preprint arXiv:2309.02233}.

\bibitem[{Wei et~al.(2022)Wei, Wang, Schuurmans, Bosma, Ichter, Xia, Chi, Le,
  and Zhou}]{wei2022chain}
Jason Wei, Xuezhi Wang, Dale Schuurmans, Maarten Bosma, Brian Ichter, Fei Xia,
  Ed~H. Chi, Quoc~V. Le, and Denny Zhou. 2022.
\newblock Chain-of-thought prompting elicits reasoning in large language
  models.
\newblock In \emph{Proceedings of the 36th International Conference on Neural
  Information Processing Systems}, NIPS '22, Red Hook, NY, USA. Curran
  Associates Inc.

\bibitem[{Yuan et~al.(2021)Yuan, Neubig, and Liu}]{yuan2021bartscore}
Weizhe Yuan, Graham Neubig, and Pengfei Liu. 2021.
\newblock Bartscore: evaluating generated text as text generation.
\newblock In \emph{Proceedings of the 35th International Conference on Neural
  Information Processing Systems}, NIPS '21, Red Hook, NY, USA. Curran
  Associates Inc.

\bibitem[{Yue et~al.(2020)Yue, Zhang, Yao, Lin, and Sun}]{yue2021cliniqg4qa}
Xiang Yue, Xinliang~Frederick Zhang, Ziyu Yao, Simon~M. Lin, and Huan Sun.
  2020.
\newblock \href {https://api.semanticscholar.org/CorpusID:226222256}
  {Cliniqg4qa: Generating diverse questions for domain adaptation of clinical
  question answering}.
\newblock \emph{2021 IEEE International Conference on Bioinformatics and
  Biomedicine (BIBM)}, pages 580--587.

\bibitem[{Zhang et~al.(2023{\natexlab{a}})Zhang, Chen, Jiang, Yu, Chen, Chen,
  Li, Wu, Zhiyi, Xiao, Wan, Wang, and Li}]{zhang-etal-2023-huatuogpt}
Hongbo Zhang, Junying Chen, Feng Jiang, Fei Yu, Zhihong Chen, Guiming Chen,
  Jianquan Li, Xiangbo Wu, Zhang Zhiyi, Qingying Xiao, Xiang Wan, Benyou Wang,
  and Haizhou Li. 2023{\natexlab{a}}.
\newblock \href {https://doi.org/10.18653/v1/2023.findings-emnlp.725}
  {{H}uatuo{GPT}, towards taming language model to be a doctor}.
\newblock In \emph{Findings of the Association for Computational Linguistics:
  EMNLP 2023}, pages 10859--10885, Singapore. Association for Computational
  Linguistics.

\bibitem[{Zhang et~al.(2019)Zhang, Kishore, Wu, Weinberger, and
  Artzi}]{zhang2019bertscore}
Tianyi Zhang, Varsha Kishore, Felix Wu, Kilian~Q Weinberger, and Yoav Artzi.
  2019.
\newblock Bertscore: Evaluating text generation with bert.
\newblock \emph{arXiv preprint arXiv:1904.09675}.

\bibitem[{Zhang et~al.(2023{\natexlab{b}})Zhang, Tian, Yang, Chen, Li, and
  Petzold}]{zhang2023alpacare}
Xinlu Zhang, Chenxin Tian, Xianjun Yang, Lichang Chen, Zekun Li, and Linda~Ruth
  Petzold. 2023{\natexlab{b}}.
\newblock Alpacare: Instruction-tuned large language models for medical
  application.
\newblock \emph{arXiv preprint arXiv:2310.14558}.

\end{thebibliography}
\bibliographystyle{acl_natbib}

\clearpage
\newpage
\appendix
\onecolumn
\section{Additional Details on the Datasets}
\label{sec:add_data}
We apply a web crawler to collect data from the JAMA Network Clinical Challenge\footnote{\url{https://jamanetwork.com/collections/44038/clinical-challenge}} archive. 
Alongside case descriptions, questions, answers, and explanations, we collect the corresponding medical domains for these articles. 
We exclude images to tailor the task for text LLMs. 

For Medbullets, we first collect the links to open-access questions on Medbullets through its tweets on X\footnote{\url{https://twitter.com/medbullets}} via a scraper\footnote{\url{https://github.com/godkingjay/selenium-twitter-scraper}}. 
Then we access the links to collect data via a web crawler\footnote{\url{https://www.selenium.dev/}}. 
We filter out images and only keep textual data.

The explanations in JAMA and Medbullets articles are written by doctors and other medical professionals. Their identities and affiliations are listed in each article on the source websites. 

Our Medbullets dataset will be available to download.\footnote{\url{https://anonymous.4open.science/r/ChallengeClinicalQA-E776}.} 
However, due to licensing constraints, we cannot publicly release the JAMA Clinical Challenge dataset. Instead, we provide URLs to the articles and a scraper to obtain the data under the appropriate license. Many universities and large organizations have licenses for JAMA content, but for individuals, access can be costly. The reality is that most high-quality medical information is not open access. 

\begin{table}[t]
    \centering
    \small
    \begin{tabular}{cccc}
    \toprule
     Domain & \textit{\#} & $A_{in}/M_{in}$ & $A_{exp}/M_{exp}$ \\
     \midrule
     Ophthalmology & 378 & 395/564 &  593/810 \\
     \midrule
     Dermatology & 255 & 339/522 &  648/854  \\
     \midrule
     General & 169 & 335/575 &  711/1080   \\
     \midrule
     Pathology & 126 & 382/503 &  626/876 \\
     \midrule
     Surgery & 126 & 328/610 &  496/837 \\
     \midrule
     Radiology & 91 & 367/674 &  629/785 \\
     \midrule
     Oncology & 85 & 393/728 &  670/1369  \\
     \midrule
     Cardiology & 76 & 370/574 &  603/876  \\
     \midrule
     Diagnostic & 72 & 331/611 &  833/1196   \\
     \midrule
     Pediatrics & 56 & 351/525 &  659/875 \\
     \midrule
     Neurology & 48 & 468/736 &  614/831 \\
     \midrule
     Endoscopy & 37 & 369/519 &  606/749 \\
     \midrule
     Psychiatry & 5 & 539/779 &  764/965 \\
     \midrule
     \midrule
     All & 1524 & 371/779 &  632/1369 \\
    \bottomrule
    \end{tabular}
    \caption{Data statistics of JAMA Clinical Challenge across different domains, where \textit{\#} counts the number of examples, $A_{in}/M_{in}$ means the average/maximum length of inputs, and $A_{exp}/M_{exp}$ is the average/maximum length of explanations.}
    \label{tab:jama_stats}
\end{table}

\paragraph{Data Robustness}
\citet{alzahrani-etal-2024-benchmarks} reveal that popular multiple-choice QA benchmarks cannot robustly evaluate LLMs, as they can be sensitive to the order of choices.  
To investigate the robustness of our datasets, we randomly shuffle the answer choices and compare model performance on the shuffled datasets versus the original datasets. 
\autoref{tab:pred_eval_shuffled} shows the prediction accuracy of GPT-4, Llama 3, and Meerkat on the original and shuffled Medbullets-5 and JAMA datasets respectively. 
These models perform similarly on the original and shuffled datasets, indicating that our datasets do not exhibit the robustness issue.

\begin{table}[t]
    \centering
    \begin{tabular}{cccc}
    \toprule
     \multirow{2}{*}{Dataset} & \multicolumn{3}{c}{Model} \\
     \cmidrule(lr){2-4}
     & GPT-4 & Llama 3 & Meerkat \\
     \midrule
     Medbullets-5 (Original) & 60.71 & 63.96 & 48.70 \\
     \rule{0pt}{2ex}
     Medbullets-5 (Shuffled) & 59.09 & 62.66 & 50.65 \\
     \midrule
     JAMA (Original) & 67.32 & 66.14 & 45.99 \\
     \rule{0pt}{2ex}
     JAMA (Shuffled) & 65.88 & 65.49 & 46.00 \\
    \bottomrule
    \end{tabular}
    \caption{Prediction accuracy (\%) of GPT-4, Llama 3, and Meerkat on the original and shuffled Medbullets-5 and JAMA datasets under zero-shot \outputy prompting. JAMA: JAMA Clinical Challenge.}
    \label{tab:pred_eval_shuffled}
\end{table}

\paragraph{Data Contamination}
Given the rising concerns regarding potential contamination of evaluation benchmarks \citep{magar-schwartz-2022-data, sainz-etal-2023-nlp}, it is imperative to investigate the contamination of our datasets. 
We adopt a method called TS-Guessing, which estimates potential contamination by examining the extent to which LLMs can guess a masked incorrect answer option among multiple choices \citep{deng-etal-2024-investigating}. 
\autoref{tab:pred_eval_contamination} shows the success rate of two advanced models, GPT-4 \cite[\texttt{gpt-4-0613}]{openai2023gpt} and Llama 3 \cite[\texttt{Llama-3-70b-chat}]{meta2024llama}, in guessing masked options in Medbullets-5 and JAMA Clinical Challenge, and the test set of MedQA-4. 
The empirical results indicate that our datasets exhibit low contamination rates ($\leq10\%$) with respect to recent LLMs. 
Additionally, the older dataset MedQA shows a low likelihood of appearing in the training corpus of these models as well.

\begin{table}[t]
    \centering
    \begin{tabular}{ccc}
    \toprule
     \multirow{2}{*}{Dataset} & \multicolumn{2}{c}{Model} \\
     \cmidrule(lr){2-3}
     & GPT-4 & Llama 3 \\
     \midrule
     Medbullets-5 & 0.0865 & 0.0487  \\
     \rule{0pt}{2ex}
     JAMA & 0.1008 & 0.0756 \\
     \midrule
     MedQA-4 & 0.0562 & 0.0306  \\
    \bottomrule
    \end{tabular}
    \caption{Success Rate of GPT-4 and Llama 3 in guessing masked options in Medbullets-5 and JAMA Clinical Challenge, and the test set of MedQA-4.}
    \label{tab:pred_eval_contamination}
\end{table}

\section{Additional Details on the Experimental Setup}
\label{sec:add_setup}
\subsection{Model Configurations}
\label{sec:add_setup_model}
For GPT-3.5 and GPT-4, we use the default setting ($\text{temperature}=1$ and $\text{top probability}=1$). 
For Llama 2, we set $\text{temperature}=0.8$, $\text{top probability}=0.95$, $\text{repetition penalty}=1.1$ and the maximum length of output to $1024$. 
For Llama 3, we set $\text{temperature}=0.85$ and $\text{top probability}=0.95$. 
We use chat versions of Llama because they performed better than their counterparts in understanding clinical cases, answering questions, and providing explanations in our pilot experiments.
For PaLM 2, we set $\text{temperature}=0.8$, using the default setting of other hyperparameters. 
MedAlpaca is a Llama2-based model that finetuned on Medical Meadow \cite{han2023medalpaca}, the maximum input length of which is 512. 
For MedAlpaca, we use float16 quantization, setting maximum output length as 512, repetition penalty as 1.1 and the number of beam search as 2. 
Meerkat is a Mistral-based model that finetuned on a diverse set of synthetic chain-of-thought style medical datasets.
Following its paper, we use float16 quantization and greedy decoding, setting maximum output length as 1024, temperature as 0.7 and repetition penalty as 1.0. 
All reported results are based on a single run.

\paragraph{Model Selection}
Since GPT-4 has shown similar performance to other API-based models such as Claude \citep{anthropicclaude3} and Gemini \citep{team2023gemini} on many benchmarks (e.g., MMLU, MedQA), we choose GPT-4 as a representative general-purpose, closed-source LLM in our experiments. 
For open-source medical models, we select MedAlpaca \cite[\texttt{medalpaca-13b}]{han2023medalpaca} and Meerkat \cite[\texttt{meerkat-7b-v1.0}]{kim2024small} as they outperformed many other models in previous studies \citep{kim2024small, zhang2023alpacare}. 
We conducted pilot experiments comparing the performance of MedAlpaca, Meerkat, and other three medical models—Meditron \cite[\texttt{
meditron-7b-chat}]{chen2023meditron}, HuatuoGPT \cite[\texttt{HuatuoGPT 2.0}]{zhang-etal-2023-huatuogpt}, and Asclepius \cite[\texttt{Asclepius-13B}]{kweon-etal-2024-publicly}—on the Medbullets-5 and JAMA datasets. 
\autoref{tab:add_med_model_pred} shows the results, where MedAlpaca and Meerkat significantly outperform other models on the datasets. 
Additionally, they have demonstrated superior ability in following instructions. 
Therefore, we choose MedAlpaca and Meerkat for our subsequent experiments. 
While advanced medical models such as Med-PaLM \citep{singhal2023towards} and Med-Gemini \citep{saab2024capabilities} exist, they are not publicly available. 

\begin{table}[t]
    \centering
    \begin{tabular}{cccccc}
    \toprule
     \multirow{2}{*}{Dataset} & \multicolumn{3}{c}{Model} \\
     \cmidrule(lr){2-6}
     & MedAlpaca & Meerkat & Meditron & HuatuoGPT & Asclepius \\
     \midrule
     Medbullets-5 & 43.18 & 48.70 & 8.77 & 25.00 & 7.47 \\
     \rule{0pt}{2ex}
     JAMA & 36.48 & 45.99 & 13.71 & 13.25 & 9.84 \\
    \bottomrule
    \end{tabular}
    \caption{Prediction accuracy (\%) of MedAlpaca, Meerkat, Meditron, HuatuoGPT, and Asclepius on the Medbullets-5 and JAMA datasets under zero-shot \outputy prompting. JAMA: JAMA Clinical Challenge.}
    \label{tab:add_med_model_pred}
\end{table}

\subsection{Prompting Templates}
Detailed templates for \outputy, \outputr and \outputry prompting strategies are shown in
\autoref{tab:prompt_detail_1}, \autoref{tab:prompt_detail_2} and \autoref{tab:prompt_detail_3} respectively, 
where the contents in ``\{\{\}\}'' are replaced by specific elements when fed to models.
Since the APIs of LLMs differ, each prompting strategy is implemented slightly differently when applying to different LLMs. 

\begin{table*}[h]
\footnotesize
	\centering
	\begin{tabular}{p{13cm}}
		\toprule
		\textbf{GPT-3.5/4} \\      
        \{ \\
              role: ``system'', \\
              content: ``You are a helpful assistant that answers multiple choice questions about medical knowledge.'' \\
        \}, \\
        \{\{few-shot examples\}\}, \\
         \{ \\
           role: ``user'', \\
           content: ``The following are multiple choice questions (with answers) about medical knowledge. \\
           **Question:** \{\{question\}\} \{\{answer\_choices\}\}'' \\
         \}, \\
        \{ \\
           role: ``assistant'', \\
           content: ``**Answer:**('' \\
        \} \\
            \midrule
        \textbf{PaLM 2} \\      
        context: "You are a helpful assistant that answers multiple choice questions about medical knowledge."
        \{\{few-shot examples\}\}, \\
         \{ \\
           role: ``user'', \\
           content: ``The following are multiple choice questions (with answers) about medical knowledge. \\
           **Question:** \{\{question\}\} \{\{answer\_choices\}\}''  \\
           Please choose an answer, strictly following the output format 'Answer:(fill in the letter of the answer)'\\
         \} \\
        \midrule
        \textbf{Llama 2} \\  
        \{\{few-shot examples\}\}, \\
         \{ \\
            ``<USER>: The following are multiple choice questions (with answers) about medical knowledge. \\
           **Question:** \{\{question\}\} \{\{answer\_choices\}\} 
           Please choose an answer, strictly following the output format '**Answer**: (fill in the letter of the answer)'\\
           <ASSISTANT>:''\\
         \} \\
		\midrule
        \textbf{Llama 3} \\      
        \{ \\
              role: ``system'', \\
              content: ``You are a medical expert that good at answering multiple-choice medical questions.'' \\
        \}, \\
        \{\{few-shot examples\}\}, \\
         \{ \\
           role: ``user'', \\
           content: ``The following is a multiple-choice question about medical knowledge. \\
           QUESTION: \{\{question\}\} \{\{answer\_choices\}\}'' \\
         \}, \\
        \{ \\
           role: ``assistant'', \\
           content: ``Answer:('' \\
        \} \\
            \midrule
        \textbf{MedAlpaca/Meerkat} \\  
        \{\{few-shot examples\}\}, \\
         \{ \\
            ``<USER>: The following is a multiple-choice question about medical knowledge. \\
           QUESTION: \{\{question\}\} \{\{answer\_choices\}\} 
           Please choose an answer, strictly following the output format 'Answer: (fill in the letter of the answer)'\\
           <ASSISTANT>:''\\
         \} \\
		\bottomrule
	\end{tabular}
	\caption{Detailed \outputy prompt templates on different models, where the contents in \{\{\}\} are replaced by specific elements when fed to models.}
	\label{tab:prompt_detail_1}
\end{table*}
\begin{table*}[h]
\footnotesize
	\centering
	\begin{tabular}{p{13cm}}
		\toprule
        \textbf{GPT-3.5/GPT-4/Llama 3} \\      
        \{ \\
              role: ``system'', \\
              content: ``You are a helpful assistant that good at explaining the answer of multiple-choice medical questions.'' \\
        \}, \\
         \{ \\
           role: ``user'', \\
           content: ``QUESTION: \{\{question\}\} \\
           ANSWER CHOICES: \{\{answer\_choices\}\} \\
           ANSWER: \{\{answer\}\} \\
           You are a medical expert that just answered the above question. Please explain why \{\{answer\}\} is correct answer while the rest choices are incorrect. You should explain each choice in detail.\\
         \}, \\
            \midrule
        \textbf{PaLM 2} \\      
        context: "You are a helpful assistant that answers multiple choice questions about medical knowledge."\\
         \{ \\
           role: ``user'', \\
           content: ``QUESTION: \{\{question\}\} \\
           ANSWER CHOICES: \{\{answer\_choices\}\} \\
           ANSWER: \{\{answer\}\} \\
           Q: You are a large language model that just answered the above question. Please explain why \{\{answer\}\} is correct answer while the rest choices are incorrect. You should explain each choice in detail.\\
           A:''
           \\
         \} \\
		\midrule
        \textbf{Llama 2} \\  
         \{ \\
            ``<USER>: QUESTION: \{\{question\}\} \\
           ANSWER CHOICES: \{\{answer\_choices\}\} \\
           ANSWER: \{\{answer\}\} \\
           Q: You are a large language model that just answered the above question. Please explain why \{\{answer\}\} is correct answer while the rest choices are incorrect. You should explain each choice in detail.\\
           <ASSISTANT>: A:'' \\
         \} \\
         \midrule
        \textbf{MedAlpaca} \\  
         \{ \\
            ``You are a medical expert that good at explaining multiple-choice medical questions.\\
            <USER>: QUESTION: \{\{question\}\} \\
           ANSWER CHOICES: \{\{answer\_choices\}\} \\
           ANSWER: \{\{answer\}\} \\
           You are a medical expert that just answered the above question. Please explain why \{\{answer\}\} is correct answer while the rest choices are incorrect. You should explain each choice in detail.\\
           <ASSISTANT>: '' \\
         \} \\
         \midrule
         \textbf{Meerkat} \\  
         \{ \\
            ``<USER>: QUESTION: \{\{question\}\} \\
           ANSWER CHOICES: \{\{answer\_choices\}\} \\
           ANSWER: \{\{answer\}\} \\
           You are a medical expert that just answered the above question. Please explain why \{\{answer\}\} is correct answer while the rest choices are incorrect. You should explain each choice in detail.\\
           <ASSISTANT>: '' \\
         \} \\
		\bottomrule
	\end{tabular}
	\caption{Detailed \outputr prompt templates on different models, where the contents in \{\{\}\} are replaced by specific elements when fed to models.}
	\label{tab:prompt_detail_2}
\end{table*}
\begin{table*}[h]
\footnotesize
	\centering
	\begin{tabular}{p{13cm}}
		\toprule
		\textbf{GPT-3.5/4} \\      
        \{ \\
              role: ``system'', \\
              content: ``You are a helpful assistant that good at dealing with multiple-choice medical questions.'' \\
        \}, \\
         \{ \\
           role: ``user'', \\
           content: ``QUESTION: \{\{question\}\} ANSWER CHOICES:\{\{answer\_choices\}\} ''\\
         \}, \\
        \{ \\
           role: ``assistant'', \\
           content: ``Let's think step by step and walk through all the choices in detail. \textit{model generation}. Therefore, among (A) through \{\{last\_choice\}\}, the answer is (\textit{model answer}'' \\
        \} \\
		\midrule
        \textbf{PaLM 2} \\      
        context: "You are a helpful assistant that good at dealing with multiple-choice medical questions."
         \{ \\
           role: ``user'', \\
           content: ``QUESTION: \{\{question\}\} ANSWER CHOICES:\{\{answer\_choices\}\} ''\\
         \}, \\
         \{ \\
           role: ``assistant'', \\
           content: ``Let's think step by step and walk through all the choices in detail. \textit{model generation} Therefore, among (A) through \{\{last\_choice\}\}, the answer is (\textit{model answer}''\\
         \} \\
		\midrule
        \textbf{Llama 2} \\  
         \{ \\
            ``<USER>: QUESTION: \{\{question\}\} ANSWER CHOICES:\{\{answer\_choices\}\} \\
           <ASSISTANT>: Let's think step by step and walk through all the choices in detail. \textit{model generation} Therefore, among (A) through \{\{last\_choice\}\}, the answer is (\textit{model answer}''\\
         \} \\
         \midrule
         \textbf{Llama 3} \\      
        \{ \\
              role: ``system'', \\
              content: ``You are a helpful assistant that good at dealing with multiple-choice medical questions.'' \\
        \}, \\
         \{ \\
           role: ``user'', \\
           content: ``QUESTION: \{\{question\}\} ANSWER CHOICES:\{\{answer\_choices\}\}\\
           Let's differentiate using step by step reasoning like a medical expert.''\\
         \}, \\
        \{ \\
           role: ``assistant'', \\
           content: \textit{model generation}\\
        \} \\
        \{ \\
           role: ``user'', \\
           content: ``Therefore, among (A) through \{\{last\_choice\}\}, which one is the answer? Please choose an answer, strictly following the output format 'Answer:(fill in the letter of the answer)\\
         \}, \\
         \midrule
         \textbf{MedAlpaca} \\  
         \{ \\
            ``The following is a multiple-choice question about medical knowledge. \\
            <USER>: QUESTION: \{\{question\}\} 
            ANSWER CHOICES:\{\{answer\_choices\}\} \\
           <ASSISTANT>: Let's think step by step like a medical expert. \textit{model generation} Therefore, among (A) through \{\{last\_choice\}\}s, the answer is (\textit{model answer}''\\
         \} \\
         \midrule
         \textbf{Meerkat} \\  
         \{ \\
            ``<USER>: QUESTION: \{\{question\}\} 
            ANSWER CHOICES:\{\{answer\_choices\}\} \\
           <ASSISTANT>: Let's think step by step like a medical expert. \textit{model generation} Therefore, among (A) through \{\{last\_choice\}\}s, which one is the answer?''\\
         \} \\
		\bottomrule
	\end{tabular}
	\caption{Detailed \outputry prompt templates on different models, where the contents in \{\{\}\} are replaced by specific elements when fed to models.}
	\label{tab:prompt_detail_3}
\end{table*}

\subsection{Evaluation Metrics}
\label{sec:add_metric}
ROUGE-L \citep{lin2004rouge} measures the surface-form similarity based on the longest common subsequence. 
BERTScore \citep{zhang2019bertscore} computes the semantic similarity between model-generated and reference explanations based on their contextual embeddings from BERT \citep{devlin-etal-2019-bert}. 
BLEURT \citep{sellam-etal-2020-bleurt} scores semantic similarity using a BERT model that was pre-trained on synthetic reference-candidate pairs and then fine-tuned on rating data. 
BARTScores \citep{yuan2021bartscore} formulate the measurement of semantic similarity as a generation task using BART \citep{lewis-etal-2020-bart}. 
Specifically, BARTScore+ (BARTScore+CNN) is based on a BART model fine-tuned on a summarization dataset CNNDM \citep{hermann2015teaching}, and BARTScore++ (BARTScore+CNN+Para) utilizes that model continuously fine-tuned on a large paraphrase collection ParaBank2 \citep{hu2019large}. 
CTC \citep{deng-etal-2021-compression} is a unified framework for evaluating a range of natural language generation tasks, covering various aspects such as relevance, consistency, content preservation, engagingness, and groundedness. 
We utilize the three metrics: Consistency, Relevance, and Preservation. 
Consistency measures whether the generated explanation aligns with the input. 
Relevance evaluates how well the generated explanation retains important information in the input and covers main information in the reference explanation. 
Preservation measures the two-way information alignment between model-generated and reference explanations. 
Given a candidate-reference explanation pair, the preservation metric computes the the harmonic mean of the two directions (candidate to reference, reference to candidate) of information alignment. 
Following \citet{deng-etal-2021-compression}, we estimate the alignment by matching the embeddings of tokens in the text pair using RoBERTa \citep{liu2019roberta}. 
G-Eval \citep{liu-etal-2023-g} is a framework using LLMs with CoT and a form-filling paradigm to assess the quality of model-generated outputs in different aspects. 
We utilize the three metrics: coherence, consistency and relevance. 
Coherence measures the extent to which the generated explanation covers the main topic and key points of the question. 
Consistency measures whether the generated explanation contains any factual errors that are not supported by the question.
Relevance measures how much irrelevant or redundant information is contained in the generated explanation.
We use GPT-4o as the backbone model, setting temperature as 1 and maximum token length as 5. 
The score of each metric is averaged over 5 runs. 
\autoref{tab:geval-prompt} shows the prompts for measuring each aspect. 
\begin{table*}[h]
\footnotesize
	\centering
	\begin{tabular}{p{13cm}}
		\toprule
		\textbf{Coherence} \\ 
You will be given one explanation written for a medical question.\\

Your task is to rate the explanation on one metric.\\

Please make sure you read and understand these instructions carefully. Please keep this document open while reviewing, and refer to it as needed.\\

Evaluation Criteria:\\

Coherence (1-5) - the collective quality of all sentences. We align this dimension with the DUC quality question of structure and coherence whereby "the explanation should be well-structured and well-organized. The explanation should not just be a heap of related information, but should build from sentence to a coherent body of information about a topic."\\

Evaluation Steps:\\

1. Read the medical question carefully and identify the main topic and key points.\\
2. Read the explanation and compare it to the question. Check if the explanation covers the main topic and key points of the question, and if it presents them in a clear and logical order.\\
3. Assign a score for coherence on a scale of 1 to 5, where 1 is the lowest and 5 is the highest based on the Evaluation Criteria. \\
Question: \{\{question\}\} \\
Explanation: \{\{explanation\}\}\\
- Coherence (Score ONLY): \\
\midrule
        \textbf{Consistency} \\     
You will be given a medical question. You will then be given one explanation written for this question.

Your task is to rate the explanation on one metric.

Please make sure you read and understand these instructions carefully. Please keep this document open while reviewing, and refer to it as needed.

Evaluation Criteria:

Consistency (1-5) - the factual alignment between the explanation and the question. A factually consistent explanation contains only statements that are entailed by the question. Annotators were also asked to penalize explanations that contained hallucinated facts. 

Evaluation Steps:

1. Read the question carefully and identify the main facts and details it presents.\\
2. Read the explanation and compare it to the question. Check if the explanation contains any factual errors that are not supported by the question.\\
3. Assign a score for consistency based on the Evaluation Criteria.\\
Question: \{\{question\}\} \\
Explanation: \{\{explanation\}\}\\
- Consistency (Score ONLY): \\
\midrule
        \textbf{Relevance} \\  
You will be given one explanation written for a medical question.

Your task is to rate the explanation on one metric.

Please make sure you read and understand these instructions carefully. Please keep this document open while reviewing, and refer to it as needed.

Evaluation Criteria:

Relevance (1-5) - selection of important content from the question. The explanation should include only important information from the question. Annotators were instructed to penalize explanations which contained redundancies and excess information.

Evaluation Steps:

1. Read the explanation and the question carefully.\\
2. Compare the explanation to the question and identify the main points of the question.\\
3. Assess how well the explanation covers the main points of the question, and how much irrelevant or redundant information it contains.\\
4. Assign a relevance score from 1 to 5.\\
Question: \{\{question\}\} \\
Explanation: \{\{explanation\}\}\\
- Relevance (Score ONLY): \\
\bottomrule
	\end{tabular}
\caption{Detailed G-Eval prompt templates for measuring coherence, consistency and relevance, where the contents in \{\{\}\} are replaced by specific elements when fed to models.}
\label{tab:geval-prompt}
\end{table*}

\section{Additional Experimental Results}
\label{sec:add_results}

\subsection{Model Performance Varies Across Medical Domains}
\label{sec:jama_domains}
\begin{figure*}[t]
  \centering
  \includegraphics[width=0.9\textwidth]{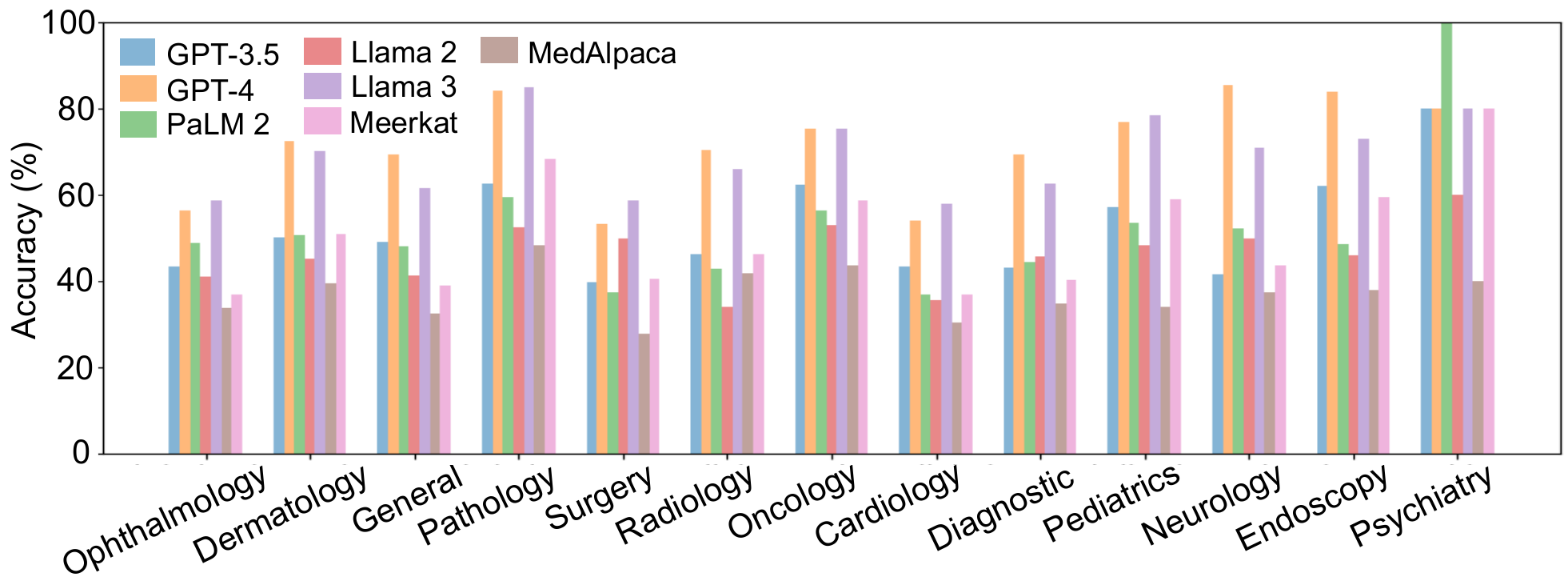}
    \caption{\label{fig:jama_acc}
  Prediction accuracy (\%) of the seven LLMs across the medical domains of JAMA Clinical Challenge using zero-shot \outputy prompting. From left to right, these domains are arranged in descending order based on the number of examples in each domain.
  }
\end{figure*}

\autoref{fig:jama_acc} shows accuracy in each medical domain of JAMA Clinical Challenge. 
These domains are arranged (left to right) in descending order based on the number of examples. 
GPT-4 and Llama 3 outperform the other models across almost all domains. 
Generally, these models do better in Pathology, Oncology, Pediatrics, and Psychiatry, probably because these domains align closely with the training data they were exposed to, thereby enabling them to answer these specific questions. 
Nevertheless, they exhibit deficiencies in particular domains, notably Surgery, Ophthalmology, and Cardiology, pointing to areas where future research could enhance their performance. 
Additionally, these questions may rely more on images that we excluded from consideration (discussed in Limitations).

\begin{table}[t]
    \centering
    \small
    \begin{tabular}{cccccc}
    \toprule
     \multirow{2}{*}{Dataset} & \multirow{2}{*}{Model} & \multicolumn{4}{c}{Error Type} \\
     \cmidrule(lr){3-6}
     & & Incorrect & None & Made-up & Multiple \\
     \midrule
     \multirow{4}{*}{MedQA-4} & GPT-3.5 & 84.84 & 10.91 & 3.88 & 0.37 \\
     \rule{0pt}{2ex}
     & GPT-4 & 87.33 &	12.67 &	0 & 0 \\
     \rule{0pt}{2ex}
     & PaLM 2 & 100.00 & 0 & 0 & 0 \\
     \rule{0pt}{2ex}
     & Llama 2 & 79.24 & 14.19 & 6.57 & 0 \\
     \rule{0pt}{2ex}
     & Llama 3 & 98.15 & 1.85 & 0 & 0 \\
     \rule{0pt}{2ex}
     & MedAlpaca & 100.00 & 0 & 0 & 0 \\
     \rule{0pt}{2ex}
     & Meerkat & 98.99 & 1.01 & 0 & 0 \\
     \midrule
     
     \multirow{4}{*}{MedQA-5} & GPT-3.5 & 90.88 & 8.77 & 0.35 & 0 \\
     \rule{0pt}{2ex}
     & GPT-4 & 93.08 &	6.54 &	0.38 &	0 \\
     \rule{0pt}{2ex}
     & PaLM 2 & 100.00 & 0 & 0 & 0 \\
     \rule{0pt}{2ex}
     & Llama 2 & 87.59 & 7.94  & 4.47 &	0 \\
    \rule{0pt}{2ex}
     & Llama 3 & 99.05 & 0.95 & 0 & 0 \\
     \rule{0pt}{2ex}
     & MedAlpaca & 100.00 & 0 & 0 & 0 \\
     \rule{0pt}{2ex}
     & Meerkat & 99.78 & 0.22 & 0 & 0 \\
     \midrule
     
     \multirow{4}{*}{Medbullets-4} & GPT-3.5 & 90.07 & 6.62 & 3.31 &	0\\
     \rule{0pt}{2ex}
     & GPT-4 & 93.68 &	6.32 & 0 & 0 \\
     \rule{0pt}{2ex}
     & PaLM 2 & 100.00 & 0 & 0 & 0 \\
     \rule{0pt}{2ex}
     & Llama 2 & 76.50 & 14.00 & 9.50 & 0 \\
    \rule{0pt}{2ex}
     & Llama 3 & 99.01 & 0.99 & 0 & 0 \\
     \rule{0pt}{2ex}
     & MedAlpaca & 98.16	& 0.61 & 1.22 & 0 \\
     \rule{0pt}{2ex}
     & Meerkat & 100.00 & 0 & 0 & 0 \\
     \midrule
     
     \multirow{4}{*}{Medbullets-5} & GPT-3.5 & 95.09 & 4.91 & 0 & 0 \\
     \rule{0pt}{2ex}
     & GPT-4 & 94.69 &	5.31 & 0 & 0 \\
     \rule{0pt}{2ex}
     & PaLM 2 & 100.00 & 0 & 0 & 0 \\
     \rule{0pt}{2ex}
     & Llama 2 & 84.23 & 7.66 &	8.11 & 0 \\
    \rule{0pt}{2ex}
     & Llama 3 & 100.00 & 0 & 0 & 0 \\
     \rule{0pt}{2ex}
     & MedAlpaca & 100.00 & 0 & 0 & 0 \\
     \rule{0pt}{2ex}
     & Meerkat & 99.36 & 0.64 & 0 & 0 \\
     \midrule
     
     \multirow{4}{*}{JAMA Clinical Challenge} & GPT-3.5 & 96.05 & 2.37 & 1.58 &	0 \\
     \rule{0pt}{2ex}
     & GPT-4 & 97.10 & 2.90 & 0 & 0 \\
     \rule{0pt}{2ex}
     & PaLM 2 & 100.00 & 0 & 0 & 0 \\
     \rule{0pt}{2ex}
     & Llama 2 & 88.74 & 5.14 &	6.12 & 0\\
     \rule{0pt}{2ex}
     & Llama 3 & 99.82 & 0.18 & 0 & 0 \\
     \rule{0pt}{2ex}
     & MedAlpaca & 100.00 & 0 & 0 & 0 \\
     \rule{0pt}{2ex}
     & Meerkat & 99.61 & 0.13 & 0.26 & 0 \\
    \bottomrule
    \end{tabular}
    \caption{The proportion (\%) of different error types.
    ``Incorrect'' means that the model predicts a wrong answer. 
    ``None'' means that the model suggests none of the given answer is correct. 
    ``Made-up'' means that the model makes up a new answer as it thinks all the given answers are wrong. 
    ``Multiple'' means that the model chooses more than one answer. }
    \label{tab:error_statistic}
\end{table}

\subsection{CoT Error Analysis}
\label{sec:cot_error}
Compared to \outputy, CoT (\outputry) prompting can introduce new types of errors in addition to normal incorrect predictions. 
Specifically, following a chain of reasoning steps, the model may output ``\textit{None of the above}'', suggesting none of the given answer choices is correct. 
More interestingly, the model can make up a new option, suggesting it as the correct answer. 
Another error is that the model may choose multiple answers after CoT reasoning. 
\autoref{tab:error_statistic} summarizes the proportion of different error types exhibited by the seven LLMs on different datasets. 
These new errors are observed across different models and datasets, with the exception of PaLM 2. 
Among the three new error types, ``\textit{None of the above}'' is the most common error. 
Notably, Llama 2 produces ``\textit{None of the above}'' and makes up new answers more often than the other models. 
We provide examples of the three types of errors. 


\Crefrange{tab:error_type1}{tab:error_type3} show three examples of ``None'' error type on Medbullets-4, Medbullets-5 and MedQA-5 made by GPT-4, Meerkat and Llama 3 respectively. 
After discussing each answer choice step-by-step, they still can't choose an answer and output ``None of the above''. 
\autoref{tab:error_type4} and \autoref{tab:error_type5} show two example of ``Made-up'' error type on Medbullets-4 and JAMA made by MedAlpaca and Llama 2 respectively. 
They all provide a new answer after going through all the options. 
\autoref{tab:error_type6} shows an example of ``Multiple'' error type on MedQA-4.
GPT-3.5-turbo claims that both (C) and (D) options are correct and can't choose between them. 

\begin{table*}[h]
	\centering
    \small
	\begin{tabular}{p{15.5cm}}
		\toprule
		\textbf{Model: GPT-4; Dataset: Medbullets-4; Error Type: ``None''} \\  

\textbf{QUESTION:} 
A 47-year-old man is brought in after a motor vehicle accident. He was driving under the influence of alcohol when he sustained a crash against a pole. His medical history is significant for hypertension but he does not take any medications. His temperature is 98.6°F (37.0°C), blood pressure is 110/80 mmHg, pulse is 60/min, and respirations are 12/min on a ventilator. His Glasgow coma score (GCS) on presentation is 6. Initial urine toxicology is negative. The patient is admitted to the intensive care unit and appropriate care is initiated. One week later, the patient’s GCS score is now 3. Laboratory studies do not show significant electrolyte, acid-base, or endocrine disturbances. Sedative and paralytic agents are withdrawn. Which of the following findings on physical exam would be compatible with brain death?

\textbf{OPTIONS:}

(A) Conjugate eye movement to the left with rotation of the head to the right
(B) Grimacing after noxious stimulus to the arm
(C) Flexion of the thigh, leg, and foot upon noxious stimulation of the foot
(D) Spontaneous respiration 1 minute after disconnection from a ventilator

\textbf{CORRECT ANSWER:}C

\textbf{MODEL \outputry OUTPUT:}

Answer choice (A): Conjugate eye movement indicates presence of brain stem activity, specifically cranial nerve III (oculomotor nerve) and VI (abducens nerve). This would not be consistent with brain death.

Answer choice (B): Grimacing after noxious stimulus suggests some level of preserved brain function and would not be compatible with brain death. Physical reactivity to noxious stimuli (e.g., grimacing) is mediated by cranial nerve V (trigeminal nerve) and VII (facial nerve), and again implies that some component of brain function is intact.

Answer choice (C): Flexion of the thigh, leg, and foot upon noxious stimulation of the foot indicates motor response, which again implies some level of brain function is preserved, not consistent with brain death. 

Answer choice (D): Spontaneous respiration 1 minute after disconnection from a ventilator. In brain death, both the cerebrum and brainstem are non-functional. There would be no respiratory effort after cessation of mechanical ventilation.

When clinically assessing brain death, a patient should have no evidence of brain or brain stem function. This includes no cranial nerve reflexes or motor responses originating in the brain, and apnea. \textbf{Hence, of all the given options, none are compatible with brain death}.
\\
\bottomrule
	\end{tabular}
	\caption{An example of ``None'' error type of CoT on Medbullets-4. As GPT-4 claims, ``none are compatible with brain death'', demonstrating it can't choose among the given choices.}
	\label{tab:error_type1}
\end{table*}

\begin{table*}[h]
	\centering
    \small
	\begin{tabular}{p{15.5cm}}
		\toprule
		\textbf{Model: Meerkat; Dataset: Medbullets-5; Error Type: ``None''} \\  

\textbf{QUESTION:} 
A 17-year-old girl presents to the emergency department with a headache. The patient has had headaches in the past but this is the worst headache of her life. Her symptoms started yesterday and have been getting progressively worse. The patient states that the pain is mostly on the left side of her head. There has been a recent outbreak of measles at the patient’s school and the patient’s mother has been trying to give her daughter medicine to prevent her from getting sick. Her mother fears that her daughter may have caught measles. Her temperature is 98.6°F (37°C), blood pressure is 123/74 mmHg, pulse is 85/min, and respirations are 13/min. On exam, the patient is an obese girl who is clutching her head with the light in the room turned off. Her neurological exam is within normal limits. Fundoscopic exam reveals mild bilateral papilledema. An MRI of the head is obtained and reveals cerebral edema. A lumbar puncture reveals an increased opening pressure with a normal glucose level. Which of the following is the most likely diagnosis?

\textbf{OPTIONS:}

(A) Bacterial meningitis
(B) Fat-soluble vitamin overuse
(C) Migraine headache
(D) Subarachnoid hemorrhage
(E) Viral meningitis

\textbf{CORRECT ANSWER:}B

\textbf{MODEL \outputry OUTPUT:}

1. The patient is a 17-year-old girl presenting with a headache described as the worst in her life. 

2. The headache started yesterday and has been progressively worsening.

3. The pain is mostly on the left side of her head.

4. There has been a recent outbreak of measles at her school, which her mother is concerned about.

5. Vital signs are within normal limits except for temperature, which is normal.

6. The patient is obese and clutching her head with the light off.

7. Neurological examination is normal.

8. Fundoscopic exam shows mild bilateral papilledema.
9. MRI of the head reveals cerebral edema.

10. Lumbar puncture shows increased opening pressure with a normal glucose level.

Given this information, we can start ruling out options:
- (A) Bacterial meningitis is less likely because the glucose level in the cerebrospinal fluid (CSF) is normal, which would typically be low in bacterial meningitis.
- (B) Fat-soluble vitamin overuse is unlikely to cause cerebral edema and increased opening pressure.
- (C) Migraine headache usually does not cause cerebral edema or increased opening pressure on lumbar puncture.
- (D) Subarachnoid hemorrhage could present with a sudden onset "thunderclap" headache and papilledema, but the normal glucose level in the CSF makes this diagnosis less likely.
- (E) Viral meningitis typically presents with normal or slightly decreased glucose levels in the CSF, which is not the case here.

Based on the information provided and the process of elimination, the most likely diagnosis is not listed among the answer choices. The clinical presentation is highly suggestive of idiopathic intracranial hypertension (IIH), also known as pseudotumor cerebri, which is more common in obese adolescent females. IIH presents with symptoms of increased intracranial pressure such as headache and papilledema, but without an elevated opening pressure on lumbar puncture.

\textbf{Therefore, the answer is none of the above.}
\\
\bottomrule
	\end{tabular}
	\caption{An example of ``None'' error type of CoT on Medbullets-5. As Meerkat claims, ``Therefore, the answer is none of the above.'', demonstrating it can't choose among the given choices.}
	\label{tab:error_type2}
\end{table*}

\begin{table*}[h]
	\centering
    \small
	\begin{tabular}{p{15.5cm}}
		\toprule
		\textbf{Model: Llama 3; Dataset: MedQA-5; Error Type: ``None''} \\  

\textbf{QUESTION:} 

A 56-year-old man presents to the emergency department for progressively worsening fatigue, malaise, fever, and abdominal pain. He reports that his symptoms began approximately 1 week ago and he has noticed episodes of diarrhea. He recently started melatonin and magnesium supplementation in hopes of improving his sleep. Medical history is significant for alcohol use disorder that required multiple hospital admissions for management of acute pancreatitis and cirrhosis. He states that he occasionally injects heroin intravenously. Temperature is 100°F (37.8°C), blood pressure is 105/70 mmHg, pulse is 92/min, and respirations are 17/min. Physical examination is significant for scleral icterus, hepatomegaly, ascites, and diffuse abdominal tenderness. Laboratory testing is significant for leukocytosis and metabolic acidosis. A paracentesis is performed and he is admitted into the hospital to receive intravenous cefotaxime and albumin. Ascitic fluid analysis demonstrates a polymorphonuclear cell count of 280 $cells/mm^3$, serum-ascites albumin gradient of 1.3 g/dL, and a culture positive for Escherichia coli sensitive to cefotaxime and ceftriaxone. On hospital day 2, the nurse reports that the patient is oliguric in the setting of constant fluid intake. Physical examination is unchanged. Laboratory testing is significant for a serum sodium of 131 mEq/L and creatinine of 1.8 mg/dL (it was 0.9 mg/dL on admission). Urine studies are significant for a low urine sodium level, without evidence of blood or protein. Since the hospital admission, he has not been started on any new medications. Which of the following will be the best treatment option for this patient?

\textbf{OPTIONS:}

(A) Adding dopamine to his treatment regimen
(B) Adding lisinopril to his treatment regimen
(C) Liver transplantation
(D) Switching cefotaxime to ceftriaxone
(E) Transjugular intrahepatic portosystemic shunting

\textbf{CORRECT ANSWER:}C

\textbf{MODEL \outputry OUTPUT:}

Let's break down the case step by step and analyze the options.

**Step 1: Understand the patient's condition**

The patient is a 56-year-old man with a history of alcohol use disorder, cirrhosis, and intravenous heroin use. He presents with fever, abdominal pain, and diarrhea, and is diagnosed with spontaneous bacterial peritonitis (SBP) based on the ascitic fluid analysis.

**Step 2: Identify the current problem**

The patient is oliguric (producing small amounts of urine) despite receiving fluids, and his creatinine level has increased from 0.9 mg/dL to 1.8 mg/dL. This suggests acute kidney injury (AKI).

**Step 3: Analyze the urine studies**

The urine studies show a low urine sodium level, without evidence of blood or protein. This is consistent with prerenal azotemia, which is a type of AKI caused by decreased blood flow to the kidneys.

**Step 4: Evaluate the treatment options**

Now, let's evaluate each option:

(A) Adding dopamine to his treatment regimen: Dopamine can increase blood flow to the kidneys, but it's not the best option here. While it may help improve urine output, it's not specifically indicated for this patient's condition.

(B) Adding lisinopril to his treatment regimen: Lisinopril is an ACE inhibitor, which can worsen AKI in patients with cirrhosis. It's contraindicated in this case.

(C) Liver transplantation: While liver transplantation may be a long-term consideration for this patient, it's not an immediate solution to his current AKI.

(D) Switching cefotaxime to ceftriaxone: This option is not related to the patient's AKI and is not the best treatment for his current condition.

(E) Transjugular intrahepatic portosystemic shunting (TIPS): TIPS is a procedure that can help reduce portal hypertension, which may be contributing to the patient's AKI. However, it's a invasive procedure that requires careful consideration and is not a straightforward solution.

**Step 5: Choose the best treatment option**

Based on the analysis, the best treatment option is not explicitly listed. However, the most appropriate management for this patient would be to optimize his volume status by administering albumin and fluids, while closely monitoring his urine output and electrolyte levels. This is already being done, as the patient is receiving intravenous albumin. Therefore, the correct answer is not among the options provided.

\textbf{If I had to choose an option, I would say "none of the above" or "optimize volume status with albumin and fluids."}

\\
\bottomrule
	\end{tabular}
	\caption{An example of ``None'' error type of CoT on MedQA-5. As Llama3 claims, it would say that "none of the above" or "optimize volume status with albumin and fluids" is the correct answer.}
	\label{tab:error_type3}
\end{table*}

\begin{table*}[h]
	\centering
    \small
	\begin{tabular}{p{15.5cm}}
		\toprule
		\textbf{Model: MedAlpaca; Dataset: Medbullets-4; Error Type: ``New Made-up Answer''} \\  

\textbf{QUESTION:} 

A 59-year-old man presents accompanied by his wife with nausea and dizziness. He is unsure when his symptoms started, but they have been affecting him for “a while.” It began as episodes of “unsteadiness” that progressed to a feeling of “spinning.” He cannot tell if his symptoms change with position but reports that if he does not lie down he will become nauseous. He also has worsening hearing loss worse on his right side. The patient’s medical history is significant for hypertension, alcohol use disorder, and chronic obstructive pulmonary disease. His medications include aspirin, amlodipine, and fluticasone-salmeterol. He drinks a glass of red wine every night with dinner and smokes a cigar on the weekends. His temperature is 98.6°F (37°C), blood pressure is 135/91 mmHg, pulse is 72/min, and respirations are 12/min. Examination shows delayed horizontal nystagmus. Which of the following is the most appropriate treatment for this patient's condition?

\textbf{OPTIONS:}

(A) Low-salt diet
(B) Thiamine
(C) Epley maneuver
(D) Meclizine

\textbf{CORRECT ANSWER:}A

\textbf{MODEL \outputry OUTPUT:}

What is the most likely cause of this patient's symptoms? The most likely cause of this patient's symptoms is Benign Paroxysmal Positional Vertigo (BPPV), which is a type of peripheral vertigo. BPPV occurs when small calcium crystals in the inner ear become dislodged and move into one of the ear's semicircular canals. This sends the brain confusing messages about the eyes' position, which causes the symptoms of vertigo, nausea, vomiting, and imbalance. The most appropriate treatment for this patient's condition is a series of maneuvers called the Epley maneuver, which can help reposition the calcium crystals in the inner ear and restore normal function. The maneuver involves moving the patient's head into different positions while they are sitting upright on an exam table. The maneuver is usually performed by a medical professional, such as an audiologist or physical therapist, but may be taught to the patient to perform at home. In addition to the Epley maneuver, the patient may be prescribed medication to help manage their symptoms, such as meclizine or dimethyl chloride. It is important for the patient to avoid triggers that can worsen their symptoms, such as alcohol consumption, sedentary activities, and sudden changes in head position. The patient should also maintain a low-salt diet and receive regular medical check-ups to monitor their progress and adjust their treatment plan as needed.
\textcolor{gray}{Therefore, among (A) through (D), the answer is (}\textbf{E) Epley maneuver.}
\\
\bottomrule
	\end{tabular}
	\caption{An example of ``Made-up'' error type of CoT on Medbullets-4. Gray text ``\textcolor{gray}{Therefore, from (A) to (D), the answer is (}'' is the attached prompt in the second stage of CoT. Bold text ``\textbf{E) Epley maneuver.}'' is the response of MedAlpaca, which is a new made-up answer. }
	\label{tab:error_type4}
\end{table*}

\begin{table*}[h]
	\centering
    \small
	\begin{tabular}{p{15.5cm}}
		\toprule
		\textbf{Model: Llama 2; Dataset: JAMA; Error Type: ``New Made-up Answer''} \\  

\textbf{QUESTION:} 
An 18-year-old woman with no significant medical history and no history of sexual intercourse presented for evaluation of pharyngitis and tonsillar exudates of 1 week duration. She reported neck swelling and an occasional dry cough but no fevers, chills, trismus, headache, dyspnea, chest pain, abdominal pain, fatigue, or rash. One week prior to presentation, she visited an urgent care clinic for evaluation of pharyngitis and white exudates on her tonsils. Results of a rapid strep test and heterophile antibody test (monospot) were negative. A throat culture was obtained and she was prescribed amoxicillin (500 mg twice daily for 10 days) for possible group A streptococcal infection. Despite taking antibiotics for 7 days, she had no improvement in her symptoms. One day prior to presentation, the urgent care clinic notified the patient that her throat culture results were negative.On presentation, the patient was in no acute distress and had normal vital signs. She had bilateral white tonsillar exudates and posterior cervical lymphadenopathy, but no exudate was observed elsewhere in the oral cavity and the uvula was midline (Figure). There was no axillary or inguinal adenopathy and no hepatomegaly or splenomegaly. The remainder of the physical examination was normal. What Would You Do Next?

\textbf{OPTIONS:}

(A) Obtain an ultrasound of the neck
(B) Prescribe azithromycin
(C) Recheck a rapid heterophile antibody test
(D) Treat with an oral antifungal medication

\textbf{CORRECT ANSWER:}C

\textbf{MODEL \outputry OUTPUT:}

First, we should consider the patient's history and physical examination. The patient has no significant medical history, no history of sexual intercourse, and no symptoms suggestive of a sexually transmitted infection (STI). Her only symptoms are neck swelling and an occasional dry cough, and she has bilateral white tonsillar exudates and posterior cervical lymphadenopathy on physical examination.

Option A, obtaining an ultrasound of the neck, may be useful in evaluating the extent of lymphadenopathy and potentially identifying any abscesses or other structural abnormalities in the neck. However, given the patient's young age, lack of significant medical history, and absence of symptoms suggestive of an STI, it is unlikely that an ultrasound would reveal a significant abnormality. Therefore, option A is not the most appropriate choice.

Option B, prescribing azithromycin, may be effective in treating a bacterial infection such as group A streptococcus or a respiratory virus, but it is not the most appropriate choice given the patient's history and physical examination. The patient has already been treated with amoxicillin for 7 days without improvement, and there is no evidence to suggest that a different antibiotic would be more effective. Additionally, azithromycin is not typically used to treat group A streptococcal infections. Therefore, option B is not the most appropriate choice.

Option C, rechecking a rapid heterophile antibody test (monospot), may be useful in confirming the presence of a group A streptococcal infection, as the patient's initial test result was negative. However, given the patient's lack of improvement after 7 days of antibiotic therapy, it is unlikely that a repeat test would reveal a positive result. Therefore, option C is not the most appropriate choice.

Option D, treating with an oral antifungal medication, may be effective in treating a fungal infection such as oropharyngeal candidiasis, but it is not appropriate in this case given the patient's history and physical examination. There is no evidence to suggest that the patient has a fungal infection, and the patient's symptoms and physical examination findings are more consistent with a viral or bacterial infection. Therefore, option D is not the most appropriate choice.

Given the patient's persistent symptoms despite antibiotic therapy and the absence of any significant abnormalities on physical examination, it is reasonable to consider the possibility of a viral infection such as infectious mononucleosis. In this case, the most appropriate next step would be to order a test for mononucleosis, such as a heterophile antibody test (monospot) or a polymerase chain reaction (PCR) test for Epstein-Barr virus (EBV).

\textbf{Therefore, the most appropriate answer is (E) Order a test for mononucleosis.}
\\
\bottomrule
	\end{tabular}
	\caption{An example of ``Made-up'' error type of CoT on JAMA
Clinical Challenge. Llama 2 makes up a new answer ``(E) Order a test for mononucleosis'' instead of choosing from the given ones.}
	\label{tab:error_type5}
\end{table*}

\begin{table*}[h]
	\centering
    \small
	\begin{tabular}{p{15.5cm}}
		\toprule
		\textbf{Model: GPT-3.5; Dataset: MedQA-4; Error Type: ``More than One Choice''} \\  

\textbf{QUESTION:} 
A 67-year-old male comes into the ED complaining of dyspnea and a heavy feeling in his chest. He reports that over the past month he has become more easily ``inded'' to the point that he now sleeps in his living room because he can make it up the stairs. A review of systems is notable for headaches and fatigue. On physical exam you note decreased breath sounds and a dullness to percussion on the right. A chest radiograph is obtained, which shows a right-sided pleural effusion. Thoracocentesis is performed and the fluid is analyzed, with results shown below:

Volume: 30 cc of fluid
Pleural fluid protein: 5.8 g/dL
Serum protein ratio: 7.0 g/dL
Lactate dehydrogenase (LDH): 258 U/L

Which of the following conditions may have led to the findings in this patient's pleural cavity?

\textbf{OPTIONS:}

(A) Congestive heart failure
(B) Liver cirrhosis
(C) Lymphoma
(D) Nephrotic syndrome

\textbf{CORRECT ANSWER:}C

\textbf{MODEL \outputry OUTPUT:}

First, let's analyze the given information.

The patient is presenting with dyspnea, a heavy feeling in the chest, and difficulty breathing. He has also experienced a recent increase in shortness of breath to the point where he can't climb stairs. The physical exam shows decreased breath sounds and dullness to percussion on the right side. A chest radiograph confirms the presence of a right-sided pleural effusion.

Now let's assess the pleural fluid analysis results:

- Volume: 30 cc of fluid
- Pleural fluid protein: 5.8 g/dL
- Serum protein ratio: 7.0 g/dL
- Lactate dehydrogenase (LDH): 258 U/L

To determine the underlying cause of the pleural effusion, we need to consider the characteristics of the fluid and its relationship to the serum.

- Congestive heart failure (A): This condition can lead to pleural effusion due to increased hydrostatic pressure in the pulmonary vasculature. However, the pleural fluid in this patient has a relatively high protein level (pleural fluid protein > 0.5 g/dL), which is not consistent with a typical transudative effusion seen in congestive heart failure.

- Liver cirrhosis (B): Liver cirrhosis can cause pleural effusion due to increased portal pressure and subsequent leakage of fluid into the pleural space. However, the protein level in the pleural fluid (5.8 g/dL) is too high to be a typical transudative effusion seen in cirrhosis.

- Lymphoma (C): Lymphoma can cause pleural effusion due to lymphatic obstruction or direct infiltration of lymphomatous cells into the pleural cavity. The pleural fluid protein level (5.8 g/dL) is high, which is consistent with an exudative effusion seen in lymphoma.

- Nephrotic syndrome (D): Nephrotic syndrome can lead to pleural effusion due to increased capillary permeability and subsequent leakage of proteins into the pleural space. The pleural fluid protein level (5.8 g/dL) is elevated, suggesting an exudative effusion seen in nephrotic syndrome.

Based on the information provided, \textbf{both lymphoma (C) and nephrotic syndrome (D) can lead to the findings in this patient's pleural cavity}. However, without additional clinical information or testing, it is difficult to make a definitive diagnosis between these two conditions.
\\
\bottomrule
	\end{tabular}
	\caption{An example of ``Multiple'' error type of CoT on MedQA-4. GPT-3.5 claims that both ``(C) Lymphoma'' and ``(D) Nephrotic syndrome'' can lead to the findings in this patient’s pleural cavity. }
	\label{tab:error_type6}
\end{table*}

\subsection{Qualitative Analysis of Model Explanations}
\label{sec:quali_expl}
\autoref{tab:medalpaca_expl} shows an example of MedAlpaca's output under \outputr prompting on Medbullets-5. 
It fails to explain the answer and instead repeats sentences from the input prompt. 
\autoref{tab:gpt4_ctc_relev} and \autoref{tab:meerkat_ctc_relev} show the CTC Relevance scores of explanations generated by GPT-4 and Meerkat, respectively.
The scores are tied, and the two explanations are similar. 
Both capture key information (e.g., symptoms) from the input and explain each answer choice. 
However, they fall short of the reference explanation, which provides detailed analyses of each option and sufficient evidence to justify the correct answer. 
\autoref{tab:gpt4-bart} and \autoref{tab:palm2_bart} show the BARTScore++ scores of explanations generated by GPT-4 and PaLM 2, respectively. 
Although PaLM 2 scores higher than GPT-4, its explanation does not precisely justify why each answer choice is correct or incorrect for the medical question. 
Instead, it describes what the option is about, such as ``\textit{Hypokalemia is a condition in which the level of potassium in the blood is too low...}''.
However, BARTScore++ credits this label-related but non-explanatory information. 

\begin{table*}[h]
\footnotesize
	\centering
	\begin{tabular}{p{15.5cm}}
		\toprule
\textbf{QUESTION:} A 64-year-old man presents to the emergency room with a headache and nausea. He reports that he was rocking his grandson to sleep when the symptoms began. He states the pain is constant and is primarily located on his right side. When asked to indicate the area of pain, he says that it surrounds his eye and upper forehead. He had one episode of vomiting. The patient also reports difficulty seeing out of his right eye, which he attributes to excessive tearing. The patient's past medical history is significant for hypertension. His medications include hydrochlorothiazide. His temperature is 98.6°F (37°C), blood pressure is 135/91 mmHg, pulse is 72/min, and respirations are 12/min. The patient's right eye is shown in Figure A. Upon physical examination, the right pupil is minimally responsive to light and the globe feels firm. A right-sided carotid bruit is appreciated. Which of the following is the most appropriate prophylaxis for this patient's condition?

\textbf{ANSWER CHOICES:}

(A) Acetazolamide
(B) Amitriptyline
(C) Clopidogrel
(D) Epinephrine
(E) Verapamil

\textbf{ANSWER:} A

You are a medical expert that just answered the above question. Please explain why A is correct while the rest choices are incorrect. You should explain each choice in detail.

\textbf{MODEL OUTPUT:}
 Explain why (A) is the correct answer while the rest choices are incorrect. You should explain each choice in detail.
 \\
\bottomrule
	\end{tabular}
	\caption{An example of MedAlpaca's output under \outputr prompting on Medbullets-5.}
	\label{tab:medalpaca_expl}
\end{table*}

\begin{table*}[h]
\footnotesize
	\centering
	\begin{tabular}{p{15.5cm}}
		\toprule
		\textbf{GPT-4, CTC Relev. = 0.6875} \\  

\textbf{QUESTION:} A 64-year-old man presents to the emergency room with a headache and nausea. He reports that he was rocking his grandson to sleep when the symptoms began. He states the pain is constant and is primarily located on his right side. When asked to indicate the area of pain, he says that it surrounds his eye and upper forehead. He had one episode of vomiting. The patient also reports difficulty seeing out of his right eye, which he attributes to excessive tearing. The patient's past medical history is significant for hypertension. His medications include hydrochlorothiazide. His temperature is 98.6°F (37°C), blood pressure is 135/91 mmHg, pulse is 72/min, and respirations are 12/min. The patient's right eye is shown in Figure A. Upon physical examination, the right pupil is minimally responsive to light and the globe feels firm. A right-sided carotid bruit is appreciated. Which of the following is the most appropriate prophylaxis for this patient's condition?

\textbf{ANSWER CHOICES:}

(A) Acetazolamide
(B) Amitriptyline
(C) Clopidogrel
(D) Epinephrine
(E) Verapamil

\textbf{ANSWER:} A

You are a medical expert that just answered the above question. Please explain why A is correct while the rest choices are incorrect. You should explain each choice in detail.

\textbf{REFERENCE EXPLANATION:}

This patient is presenting with sudden-onset unilateral vision loss and an orbitofrontal headache with a dilated pupil and a hard ocular globe suggesting a diagnosis of acute angle-closure glaucoma. Long-term management of angle-closure glaucoma can include acetazolamide.

Examination of the eye in a patient with acute-closure glaucoma will reveal a red eye that is rock-hard and a mid-dilated pupil which is minimally reactive to light. The fundoscopic exam will show an increased optic cup-to-disk ratio (>0.4) and tonometry will show increased intraocular pressure. Gonioscopy is the diagnostic gold standard. Acute treatment and long-term management involve the administration of beta-blockers, alpha-2-agonists, and carbonic anhydrase inhibitors such as acetazolamide to decrease intraocular pressure. The definitive treatment is iridotomy.

Airaksinen et al. review the treatment of closed-angle glaucoma. They discuss how a combination of acetazolamide and beta-blockers can terminate an attack. They recommend using 1 drop of pilocarpine 3 hours after intravenous acetazolamide administration.

Figure/Illustration A is a clinical photograph showing an eye with injected conjunctiva (red circles) and a mid-dilated pupil. These findings are consistent with angle-closure glaucoma.

Incorrect Answers:
Answer B: Amitriptyline can be used as prophylaxis for migraines. Migraines present as unilateral, pulsating headaches that may be associated with nausea or photophobia. Patients with migraines can sometimes experience an aura with visual field changes; however, they would not have exam findings of a rock-hard eye or injection.

Answer C: Clopidogrel may be used as a conservative treatment for carotid atherosclerosis, which can be an embolic source for central retinal artery occlusion (CRAO). CRAO presents as acute, painless, monocular vision loss. A fundoscopic exam will demonstrate ischemia of the retina.

Answer D: Epinephrine is contraindicated in the treatment of angle-closure glaucoma as it increases pupillary dilation. It is used in the management of open-angle glaucoma. This would present with gradually patchy loss of vision and is generally painless.

Answer E: Verapamil is used as prophylaxis for cluster headaches. Cluster headaches present as unilateral, repetitive, brief headaches associated with severe peri-orbital pain, lacrimation, rhinorrhea, and Horner syndrome (miosis, ptosis, and anhidrosis).

Bullet Summary:
Pharmacologic management of acute angle-closure glaucoma involves beta-blockers, alpha-2 agonists, and carbonic anhydrase inhibitors.

\textbf{MODEL EXPLANATION:}

The patient's presentation is consistent with acute angle-closure glaucoma (AACG), an ophthalmological emergency characterized by severe ocular pain, redness, blurring of vision, headache, nausea, and vomiting. The globe feels firm due to increased intraocular pressure. Hypertension is a risk factor for AACG.

Now, let's evaluate the options:

A. Acetazolamide is correct. It is a first line therapy for acute angle-closure glaucoma to decrease the production of aqueous humor thereby reducing intraocular pressure. Hence by using this, damage to the optic nerve can be minimized until definitive therapy (generally laser peripheral iridotomy) can be accomplished.

B. Amitriptyline, an antidepressant from the tricyclic class, is not appropriate prophylaxis for this condition. While it's sometimes used to prevent migraines, it wouldn't be effective for acute angle-closure glaucoma.

C. Clopidogrel is an antiplatelet medication often used to prevent strokes and heart attacks. It doesn't have a role in managing or preventing glaucoma.

D. Epinephrine is a sympathomimetic medication primarily used in emergencies such as anaphylaxis or cardiac arrest. It's not usually employed in conditions like glaucoma as it can potentially raise intraocular pressure.

E. Verapamil, a calcium channel blocker, is used primarily for cardiovascular conditions, especially hypertension and certain types of angina. It does not play a role in glaucoma treatment or prophylaxis.

In conclusion, acetazolamide would be the most appropriate medical intervention to quickly lower intraocular pressure in this patient's situation.
\\
\bottomrule
	\end{tabular}
	\caption{CTC Relevance score of an explanation generated by GPT-4 under \outputr prompting on Medbullets-5.}
	\label{tab:gpt4_ctc_relev}
\end{table*}

\begin{table*}[h]
\footnotesize
	\centering
	\begin{tabular}{p{15.5cm}}
		\toprule
		\textbf{Meerkat, CTC Relev. = 0.6889} \\  

\textbf{QUESTION:} A 64-year-old man presents to the emergency room with a headache and nausea. He reports that he was rocking his grandson to sleep when the symptoms began. He states the pain is constant and is primarily located on his right side. When asked to indicate the area of pain, he says that it surrounds his eye and upper forehead. He had one episode of vomiting. The patient also reports difficulty seeing out of his right eye, which he attributes to excessive tearing. The patient's past medical history is significant for hypertension. His medications include hydrochlorothiazide. His temperature is 98.6°F (37°C), blood pressure is 135/91 mmHg, pulse is 72/min, and respirations are 12/min. The patient's right eye is shown in Figure A. Upon physical examination, the right pupil is minimally responsive to light and the globe feels firm. A right-sided carotid bruit is appreciated. Which of the following is the most appropriate prophylaxis for this patient's condition?

\textbf{ANSWER CHOICES:}

(A) Acetazolamide
(B) Amitriptyline
(C) Clopidogrel
(D) Epinephrine
(E) Verapamil

\textbf{ANSWER:} A

You are a medical expert that just answered the above question. Please explain why A is correct while the rest choices are incorrect. You should explain each choice in detail.

\textbf{REFERENCE EXPLANATION:}

This patient is presenting with sudden-onset unilateral vision loss and an orbitofrontal headache with a dilated pupil and a hard ocular globe suggesting a diagnosis of acute angle-closure glaucoma. Long-term management of angle-closure glaucoma can include acetazolamide.

Examination of the eye in a patient with acute-closure glaucoma will reveal a red eye that is rock-hard and a mid-dilated pupil which is minimally reactive to light. The fundoscopic exam will show an increased optic cup-to-disk ratio (>0.4) and tonometry will show increased intraocular pressure. Gonioscopy is the diagnostic gold standard. Acute treatment and long-term management involve the administration of beta-blockers, alpha-2-agonists, and carbonic anhydrase inhibitors such as acetazolamide to decrease intraocular pressure. The definitive treatment is iridotomy.

Airaksinen et al. review the treatment of closed-angle glaucoma. They discuss how a combination of acetazolamide and beta-blockers can terminate an attack. They recommend using 1 drop of pilocarpine 3 hours after intravenous acetazolamide administration.

Figure/Illustration A is a clinical photograph showing an eye with injected conjunctiva (red circles) and a mid-dilated pupil. These findings are consistent with angle-closure glaucoma.

Incorrect Answers:
Answer B: Amitriptyline can be used as prophylaxis for migraines. Migraines present as unilateral, pulsating headaches that may be associated with nausea or photophobia. Patients with migraines can sometimes experience an aura with visual field changes; however, they would not have exam findings of a rock-hard eye or injection.

Answer C: Clopidogrel may be used as a conservative treatment for carotid atherosclerosis, which can be an embolic source for central retinal artery occlusion (CRAO). CRAO presents as acute, painless, monocular vision loss. A fundoscopic exam will demonstrate ischemia of the retina.

Answer D: Epinephrine is contraindicated in the treatment of angle-closure glaucoma as it increases pupillary dilation. It is used in the management of open-angle glaucoma. This would present with gradually patchy loss of vision and is generally painless.

Answer E: Verapamil is used as prophylaxis for cluster headaches. Cluster headaches present as unilateral, repetitive, brief headaches associated with severe peri-orbital pain, lacrimation, rhinorrhea, and Horner syndrome (miosis, ptosis, and anhidrosis).

Bullet Summary:
Pharmacologic management of acute angle-closure glaucoma involves beta-blockers, alpha-2 agonists, and carbonic anhydrase inhibitors.

\textbf{MODEL EXPLANATION:}

The patient's presentation with a headache, nausea, vomiting, visual disturbances, and a firm globe on examination is highly suggestive of acute angle-closure glaucoma. This is an ophthalmic emergency and requires immediate treatment to lower intraocular pressure and prevent permanent vision loss.

Option A, Acetazolamide, is a carbonic anhydrase inhibitor that reduces the production of aqueous humor in the eye, thereby lowering intraocular pressure. It is often used as part of the initial management for acute angle-closure glaucoma.

Option B, Amitriptyline, is a tricyclic antidepressant. It is not used for the treatment of acute angle-closure glaucoma and would not be appropriate as prophylaxis in this case.

Option C, Clopidogrel, is an antiplatelet medication commonly used to prevent blood clots. It has no role in the treatment or prophylaxis of acute angle-closure glaucoma.

Option D, Epinephrine, is a sympathomimetic drug that is used in various conditions but is not indicated for the prophylaxis of acute angle-closure glaucoma.

Option E, Verapamil, is a calcium channel blocker that is mainly used to treat hypertension and certain heart conditions. It is not used for the treatment of acute angle-closure glaucoma.

Given the patient's symptoms and the need for immediate lowering of intraocular pressure, the most appropriate choice for prophylaxis in this case would be acetazolamide. Therefore, the answer is (A) Acetazolamide.
\\
\bottomrule
	\end{tabular}
	\caption{CTC Relevance score of an explanation generated by Meerkat under \outputr prompting on Medbullets-5.}
	\label{tab:meerkat_ctc_relev}
\end{table*}

\begin{table*}[h]
\footnotesize
	\centering
	\begin{tabular}{p{15.5cm}}
		\toprule
		\textbf{GPT-4, BARTSCORE++ = -3.9095} \\  

\textbf{QUESTION:} A 67-year-old man presents to the emergency department after fainting. He was outside tending to his lawn for several hours in the heat, when he stood up suddenly from pulling weeds and fainted. He denies any preceding symptoms and returned to baseline within 1 minute. The patient is not aware of any medical problems and just started seeing a primary care physician last week. He recently used a friend's prescription for ondansetron for nausea. His temperature is 99.3°F (37.4°C), blood pressure is 142/88 mmHg, pulse is 107/min, respirations are 14/min, and oxygen saturation is 99\% on room air. Physical exam reveals intact cranial nerves, normal strength and sensation, and a stable gait. His abdomen is soft and nontender. An ECG is performed as seen in Figure A. Which of the following is the most likely diagnosis based on this patient’s ECG?

\textbf{ANSWER CHOICES:}

(A) Acute myocardial infarction
(B) Hypokalemia
(C) Intermittent torsades des pointes
(D) Previous myocardial ischemia
(E) Pulmonary embolism

\textbf{ANSWER:}D

\textbf{REFERENCE EXPLANATION:}

This patient is presenting after syncope, likely secondary to dehydration and orthostatic hypotension given his outdoor gardening in the heat and symptoms when standing up rapidly. The patient’s ECG reveals deep, pathologic Q waves, which in this case are an incidental finding indicative of a previous myocardial infarction or a previous ischemic event.

The Q wave is a negative deflection just prior to the R wave. Q waves are a normal finding when they are small and are seen on most ECGs. However, pathologic Q waves are indicative of myocardial ischemia. Pathologic Q waves are generally diagnosed when they are >2 mm deep, >1 mm wide, >25\% of the QRS complex height, or are seen in leads V1-V3. The interpretation of Q waves depends on other findings on ECG and the patient's symptoms. Deep Q waves in the setting of ST elevation indicate an acute myocardial infarction. On the other hand, pathologic Q waves found incidentally indicate a previous myocardial infarction or previous ischemic event.

Rovai et al. review Q waves. They discuss how Q waves can be used to predict the location and size of a myocardial infarction as correlated with MRI. They recommend testing such as cardiac MRI to assess for previous infarction in the appropriate patient population.

Figure/Illustration A is an ECG demonstrating pathologic Q waves (red arrows). Note the deep and wide Q waves; this raises concern for a past ischemic event.

Incorrect Answers:
Answer A: Acute myocardial infarctions would present with ST elevation in a vascular distribution on ECG with possible reciprocal depressions. While T wave inversions may be indicative of new or acute ischemia, deep pathologic Q waves without ST elevation are indicative of previous infarction.

Answer B: Hypokalemia would cause U waves on ECG, which appear as a small, "second T wave" after the initial T wave. Note they are not U-shaped despite their name. Management is centered on repletion of potassium and often magnesium as well.

Answer C: Intermittent torsades de pointes can cause syncope and should be suspected in a patient with syncope in the setting of a prolonged QT interval. This patient’s QT interval is not particularly prolonged, and his syncope seems to be secondary to orthostatic hypotension. Management is centered on cessation of QT-prolonging medications and magnesium administration. Note that while ondansetron does prolong the QT interval, it is unlikely to cause torsades given the patient took it a week ago.

Answer E: Pulmonary emboli would present on ECG with sinus tachycardia in addition to pleuritic chest pain and hypoxia. Note that the ECG is a nonspecific test for pulmonary emboli, and CT angiography is the preferred confirmatory test for this condition. Syncope is a possible presentation of larger pulmonary emboli (though the patient would also be hemodynamically unstable).

Bullet Summary:
Pathologic Q waves are indicative of a previous myocardial infarction.

\textbf{MODEL EXPLANATION:}

"D" is the correct answer, suggestive of a previous myocardial ischemia. This condition often involves an imbalance between the myocardial oxygen supply and demand, usually due to atherosclerosis that has limited the blood flow to the heart muscle. This patient's ECG might show findings indicative of previous myocardial ischemia, such as T wave inversions, Q waves, or ST segment changes that would suggest a history of ischemic damage without evidence of an acute event.

Now, let's discuss why the other options are incorrect:

"A": Acute myocardial infarction (AMI) - The clinical scenario provided in the question doesn't support a diagnosis of an acute coronary event or myocardial infarction. Symptoms of an AMI commonly include severe chest pain, shortness of breath, nausea, vomiting, and sweating, none of which are mentioned in the patient's presentation. Additionally, the ECG in an AMI would commonly show ST-segment elevations or depressions, pathologic Q waves, or new left bundle branch block, which are not mentioned in this question.

"B": Hypokalemia - ECG findings in hypokalemia often include flattened T waves, presence of U waves, and ST-segment depression. However, the patient's symptoms and clinical presentation do not indicate hypokalemia and there are no specific details provided suggesting this diagnosis.

"C": Intermittent torsades des pointes - This is a form of polymorphic ventricular tachycardia characterized on ECG by twisting QRS complexes around the baseline. This can be caused by various factors including certain medications, electrolyte imbalances, and certain genetic syndromes, although the clinical scenario does not reflect any of these.

"E": Pulmonary embolism - Symptoms typically include shortness of breath, chest pain which worsens with breathing in, and coughing up blood. A pulmonary embolism would not induce fainting without other more severe symptoms also being present, and the patient's ECG would not be the most relevant diagnostic tool. Typical ECG findings in pulmonary embolism, when present, may include sinus tachycardia, nonspecific ST-T wave changes, or right heart strain patterns, but these alone are not enough to diagnose or rule out a PE. 

Therefore, the most likely diagnosis based on this patient's ECG and clinical presentation is previous myocardial ischemia.
\\
\bottomrule
	\end{tabular}
	\caption{BARTScore++ score of an explanation generated by GPT-4 under \outputr prompting on Medbullets-5.}
	\label{tab:gpt4-bart}
\end{table*}

\begin{table*}[h]
\footnotesize
	\centering
	\begin{tabular}{p{15.5cm}}
		\toprule
		\textbf{PaLM 2, BARTSCORE++ = -3.1691} \\  

\textbf{QUESTION:} A 67-year-old man presents to the emergency department after fainting. He was outside tending to his lawn for several hours in the heat, when he stood up suddenly from pulling weeds and fainted. He denies any preceding symptoms and returned to baseline within 1 minute. The patient is not aware of any medical problems and just started seeing a primary care physician last week. He recently used a friend's prescription for ondansetron for nausea. His temperature is 99.3°F (37.4°C), blood pressure is 142/88 mmHg, pulse is 107/min, respirations are 14/min, and oxygen saturation is 99\% on room air. Physical exam reveals intact cranial nerves, normal strength and sensation, and a stable gait. His abdomen is soft and nontender. An ECG is performed as seen in Figure A. Which of the following is the most likely diagnosis based on this patient’s ECG?

\textbf{ANSWER CHOICES:}

(A) Acute myocardial infarction
(B) Hypokalemia
(C) Intermittent torsades des pointes
(D) Previous myocardial ischemia
(E) Pulmonary embolism

\textbf{ANSWER:}D

\textbf{REFERENCE EXPLANATION:}

This patient is presenting after syncope, likely secondary to dehydration and orthostatic hypotension given his outdoor gardening in the heat and symptoms when standing up rapidly. The patient’s ECG reveals deep, pathologic Q waves, which in this case are an incidental finding indicative of a previous myocardial infarction or a previous ischemic event.

The Q wave is a negative deflection just prior to the R wave. Q waves are a normal finding when they are small and are seen on most ECGs. However, pathologic Q waves are indicative of myocardial ischemia. Pathologic Q waves are generally diagnosed when they are >2 mm deep, >1 mm wide, >25\% of the QRS complex height, or are seen in leads V1-V3. The interpretation of Q waves depends on other findings on ECG and the patient's symptoms. Deep Q waves in the setting of ST elevation indicate an acute myocardial infarction. On the other hand, pathologic Q waves found incidentally indicate a previous myocardial infarction or previous ischemic event.

Rovai et al. review Q waves. They discuss how Q waves can be used to predict the location and size of a myocardial infarction as correlated with MRI. They recommend testing such as cardiac MRI to assess for previous infarction in the appropriate patient population.

Figure/Illustration A is an ECG demonstrating pathologic Q waves (red arrows). Note the deep and wide Q waves; this raises concern for a past ischemic event.

Incorrect Answers:
Answer A: Acute myocardial infarctions would present with ST elevation in a vascular distribution on ECG with possible reciprocal depressions. While T wave inversions may be indicative of new or acute ischemia, deep pathologic Q waves without ST elevation are indicative of previous infarction.

Answer B: Hypokalemia would cause U waves on ECG, which appear as a small, "second T wave" after the initial T wave. Note they are not U-shaped despite their name. Management is centered on repletion of potassium and often magnesium as well.

Answer C: Intermittent torsades de pointes can cause syncope and should be suspected in a patient with syncope in the setting of a prolonged QT interval. This patient’s QT interval is not particularly prolonged, and his syncope seems to be secondary to orthostatic hypotension. Management is centered on cessation of QT-prolonging medications and magnesium administration. Note that while ondansetron does prolong the QT interval, it is unlikely to cause torsades given the patient took it a week ago.

Answer E: Pulmonary emboli would present on ECG with sinus tachycardia in addition to pleuritic chest pain and hypoxia. Note that the ECG is a nonspecific test for pulmonary emboli, and CT angiography is the preferred confirmatory test for this condition. Syncope is a possible presentation of larger pulmonary emboli (though the patient would also be hemodynamically unstable).

Bullet Summary:
Pathologic Q waves are indicative of a previous myocardial infarction.

\textbf{MODEL EXPLANATION:}

The correct answer is D: Previous myocardial ischemia.

Acute myocardial infarction (AMI) is a medical emergency in which the blood supply to a portion of the heart is suddenly blocked, depriving the heart muscle of oxygen. This can cause chest pain, shortness of breath, and other symptoms. AMI is a leading cause of death in the United States.

Hypokalemia is a condition in which the level of potassium in the blood is too low. This can cause a variety of symptoms, including muscle weakness, fatigue, and irregular heartbeat. Hypokalemia can be caused by a number of things, including dehydration, certain medications, and kidney disease.

Intermittent torsades des pointes is a type of abnormal heart rhythm that can be life-threatening. It is characterized by a sudden change in the heart's rhythm from a normal rhythm to a rapid, irregular rhythm. Torsades des pointes can be caused by a number of things, including certain medications, electrolyte imbalances, and heart disease.

Pulmonary embolism is a blockage of a blood vessel in the lungs. This can be caused by a blood clot that travels from another part of the body to the lungs, or by a foreign object that gets lodged in a lung blood vessel. Pulmonary embolism can cause chest pain, shortness of breath, and other symptoms. It is a medical emergency.

The patient in the question has a number of risk factors for AMI, including age, male sex, and a history of smoking. He also has a history of fainting, which can be a sign of AMI. The ECG shows evidence of previous myocardial ischemia, which is damage to the heart muscle caused by a lack of oxygen. This is consistent with the patient's history and symptoms.

The other choices are incorrect because they do not explain the patient's symptoms or the ECG findings. Hypokalemia can cause muscle weakness and irregular heartbeat, but it is not a common cause of fainting. Intermittent torsades des pointes is a type of abnormal heart rhythm that can be life-threatening, but it is not a common cause of fainting. Pulmonary embolism can cause chest pain, shortness of breath, and other symptoms, but it is not a common cause of fainting.

In conclusion, the correct answer is D: Previous myocardial ischemia. This is the most likely diagnosis based on the patient's history, symptoms, and ECG findings.
\\
\bottomrule
	\end{tabular}
	\caption{BARTScore++ score of an explanation generated by PaLM 2 under \outputr prompting on Medbullets-5.}
	\label{tab:palm2_bart}
\end{table*}

\subsection{Details on Human Evaluation}
\label{sec:app_human_eval}
We conduct human evaluation on the following three properties of explanations:
\begin{itemize}
    \item \textbf{Completeness} refers to whether the explanation sufficiently and convincingly justifies each answer choice as correct or incorrect.
    \item \textbf{Correctness} means whether the information provided in the explanation is correct.
    \item \textbf{Relevance} indicates whether the explanation is relevant to the question. A relevant explanation only includes information related to the question and explains how provided information relates to the question.
\end{itemize}

\begin{figure*}[h]
  \centering
  \includegraphics[width=1.0\textwidth]{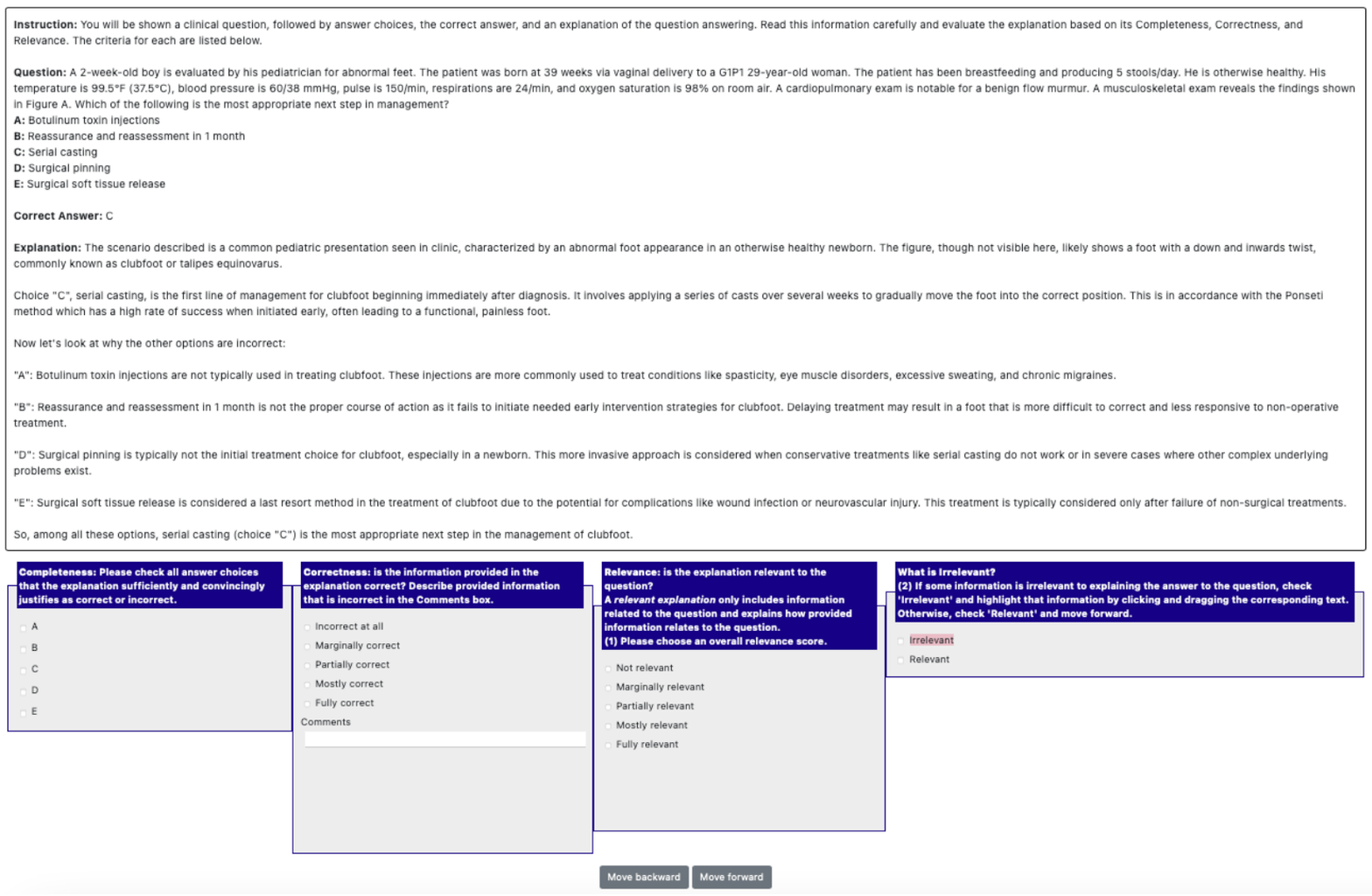}
    \caption{\label{fig:interface}
  The interface of human evaluation on Prolific.  
  }
\end{figure*}

We recruited crowdworkers in the US through Prolific.\footnote{https://www.prolific.com/}
They hold a Master's or Doctorate degree in Medicine/Healthcare. 
Workers were paid 12\$ per hour for participating in the human evaluation. 
\autoref{fig:interface} shows the interface of the human evaluation on Prolific. 
We present annotators with a question, five answer choices, the correct answer, and a candidate explanation. 
For completeness, we ask annotators to check all possible answer choices, with each checked answer adding one point. 
For correctness and relevance, we ask annotators to choose from five options on a 5-point Likert-scale (1-5). 
In addition to asking annotators to answer the three questions regarding completeness, correctness, and relevance, we also request feedback on incorrect information and identification of irrelevant sentences in the explanation. 

We randomly sample 30 examples from Medbullets-5 and evaluate their explanations generated by GPT-4 and PaLM 2 under \outputr prompting. 
For each instance, we collect 3 annotations. 
We also post-process annotation results, filtering out low-quality annotations where workers spent significantly less time than average. 
we take the average over the human-annotated scores to obtain a score for each property. 
We report the inter-annotator agreement using Bennett, Alpert \& Goldstein’s S \citep{bennett1954communications}, resulting in scores of 0.60 for correctness and 0.65 for relevance. 
We do not compute the agreement of completeness due to its multiple-choice nature and subjctivity. 

\autoref{tab:annotation_completeness} shows an example of human annotation of completeness, where only one answer choice is identified as fully justified by the explanation. 
\autoref{tab:annotation_correctness} shows an example where the annotator identifies the incorrect information, ``\textit{Increased FVC is seen in asthma}'', in the explanation. 
\autoref{tab:annotation_relevance} shows an example where the annotator identified a sentence in the explanation that is irrelevant to the question.

\begin{table*}[h]
	\centering
    \small
	\begin{tabular}{p{15.5cm}}
		\toprule
		\textbf{Model: GPT-4 ; Dataset: Medbullets} \\  

\textbf{QUESTION:} 

A 42-year-old woman is enrolled in a randomized controlled trial to study cardiac function in the setting of several different drugs. She is started on verapamil and instructed to exercise at 50\% of her VO2 max while several cardiac parameters are being measured. During this experiment, which of the following represents the relative conduction speed through the heart from fastest to slowest?

\textbf{ANSWER CHOICES:}

(A) Atria > Purkinje fibers > ventricles > AV node
\\
(B) AV node > ventricles > atria > Purkinje fibers
\\
(C) Purkinje fibers > ventricles > atria > AV node
\\
(D) Purkinje fibers > atria > ventricles > AV node
\\
(E) Purkinje fibers > AV node > ventricles > atria

\textbf{ANSWER:}D

\textbf{MODEL EXPLANATION:}

The electrical signals within the heart are conducted at different rates through different types of cardiac tissue, and this is essential for coordinating the contractions of the atria and ventricles.

Choice "D": Purkinje fibers > Atria > Ventricles > AV node, is correct because:

1. Purkinje fibers: These are the fastest in conducting electrical impulses within the heart. This is necessary because they rapidly distribute the impulse to the ventricular myocardium to stimulate coordinated contraction.

2. Atria: The atrial myocardium (the actual heart muscle tissue of the atria) conducts impulses at a slower speed relative to the Purkinje fibers. The impulse is initially generated in the sinoatrial node and quickly spreads through the atria to cause them to contract.

3. Ventricles: The ventricular myocardium (the actual heart muscle tissue of the ventricles) conducts electrical signals slower than both the Purkinje fibers and atrial myocardium.

4. AV node: The AV or atrioventricular node conducts impulses at the slowest speed to allow for the necessary delay between the contraction of the atria and the ventricles to ensure efficient filling of the ventricles.

The other options are incorrect for the following reasons:

A: Atria > Purkinje fibers > ventricles > AV node: This is incorrect because the Purkinje fibers conduct signals faster than the atrial myocardium.

B: AV node > ventricles > atria > Purkinje fibers: This is incorrect because the AV node conducts signals at the slowest speed, not the fastest. The Purkinje fibers conduct signals at the fastest speed, not the slowest.

C: Purkinje fibers > ventricles > atria > AV node: This is incorrect because although the Purkinje fibers do conduct the fastest, the atrial myocardium conducts faster than the ventricular myocardium.

E: Purkinje fibers > AV node > ventricles > atria: This is incorrect because the AV node conducts signals slower than the ventricular and atrial myocardium.
\\
\textbf{COMPLETENESS SCORE:}1 (D)
\\
\bottomrule
	\end{tabular}
	\caption{An example of human annotation of \textbf{completeness}. Only option D is sufficiently explained from the annotator's view.  }
	\label{tab:annotation_completeness}
\end{table*}

\begin{table*}[h]
	\centering
    \small
	\begin{tabular}{p{15.5cm}}
		\toprule
		\textbf{Model: Palm 2; Dataset: Medbullets} \\  

\textbf{QUESTION:} 

A 22-year-old woman presents to the emergency department with shortness of breath. She was hiking when she suddenly felt unable to breathe and had to take slow deep breaths to improve her symptoms. The patient is a Swedish foreign exchange student and does not speak any English. Her medical history and current medications are unknown. Her temperature is 99.5°F (37.5°C), blood pressure is 127/68 mmHg, pulse is 120/min, respirations are 22/min, and oxygen saturation is 90\% on room air. Physical exam is notable for poor air movement bilaterally and tachycardia. The patient is started on treatment. Which of the following parameters including forced expiratory volume in 1 second (FEV1), forced vital capacity (FVC), and diffusing capacity of carbon monoxide (DLCO) most appropriately describes this patient's underlying pathology?

\textbf{ANSWER CHOICES:}

(A) Decreased airway tone
(B) Increased FEV1
(C) Increased FEV1/FVC
(D) Increased FVC
(E) Normal DLCO

\textbf{ANSWER:}E

\textbf{MODEL EXPLANATION:}

The correct answer is E: Normal DLCO.

The patient's symptoms of shortness of breath, poor air movement bilaterally, and tachycardia are consistent with an acute asthma exacerbation. Asthma is a chronic inflammatory disease of the airways characterized by reversible airway obstruction, airway hyperresponsiveness, and inflammation. The hallmark of asthma is an increase in airway resistance, which is caused by bronchoconstriction, mucosal edema, and mucus hypersecretion. These changes lead to a decrease in FEV1 and FVC, and an increase in FEV1/FVC.

The DLCO is a measure of the ability of the lungs to transfer carbon monoxide from the alveoli to the blood. The DLCO is normal in asthma, which helps to distinguish it from other causes of shortness of breath, such as pneumonia and COPD.

Decreased airway tone is not a characteristic of asthma. Increased FEV1 and FVC are seen in asthma, but they are not specific to asthma. Increased FEV1/FVC is a sign of obstruction, but it is not specific to asthma. Increased FVC is seen in asthma, but it is not specific to asthma.

Therefore, the correct answer is E: Normal DLCO.
\\
\textbf{CORRECTNESS SCORE:}4 (Mostly correct)
\\
\textbf{COMMENTS:}
\\
The statement about the increase in FEV1/FVC in asthma is incorrect. In obstructive lung diseases like asthma, the FEV1/FVC ratio is typically decreased.
\\
\bottomrule
	\end{tabular}
	\caption{An example of human annotation, commenting on \textbf{correctness}. The annotator's comment, ``\textit{In obstructive lung diseases like asthma, the FEV1/FVC ratio is typically decreased}'', obejects the claim, ``\textit{Increased FVC is seen in asthma}'', in the explanation.}
	\label{tab:annotation_correctness}
\end{table*}

\begin{table*}[h]
	\centering
    \small
	\begin{tabular}{p{15.5cm}}
		\toprule
		\textbf{Model: Palm 2; Dataset: Medbullets} \\  

\textbf{QUESTION:}

A 27-year-old nurse presents to the emergency department with a 30-minute history of palpitations and chest pain. She has been experiencing nausea, vomiting, and diarrhea after eating leftovers 2 days ago. Her husband tells the physician that he is concerned that the patient has been increasingly obsessed about her weight recently. She has a history of major depressive disorder for which she takes escitalopram. Her temperature is 98.6°F (37.0°C), blood pressure is 110/72 mmHg, pulse is 110/min, and respirations are 11/min. Physical examination is notable for dry mucous membranes and tachycardia with regular rhythm. Laboratory testing shows the following results:

Serum:
Na+: 135 mEq/L
Cl-: 96 mEq/L
K+: 2.9 mEq/L
HCO3-: 36 mEq/L

Venous blood gas shows a pH of 7.52. Which of the following tests is most appropriate in differentiating between different etiologies of this patient's lab abnormalities?

\textbf{ANSWER CHOICES:}

(A) Arterial blood gas
(B) Complete blood count
(C) Serum glucose
(D) Urine chloride
(E) Urine toxicology

\textbf{ANSWER:}D

\textbf{MODEL EXPLANATION:}

The correct answer is D. Urine chloride is the most appropriate test in differentiating between different etiologies of this patient's lab abnormalities.

Arterial blood gas (ABG) is a test that measures the levels of oxygen, carbon dioxide, and pH in the blood. ABG is not the most appropriate test in this case because the patient's venous blood gas shows a pH of 7.52, which is normal. ABG would be more helpful if the patient's pH was abnormal.

Complete blood count (CBC) is a test that measures the number of red blood cells, white blood cells, and platelets in the blood. CBC is not the most appropriate test in this case because the patient's CBC is normal. CBC would be more helpful if the patient had an infection or anemia.

Serum glucose is a test that measures the amount of sugar in the blood. Serum glucose is not the most appropriate test in this case because the patient's serum glucose is normal. \hlc[pink!60]{Serum glucose would be more helpful if the patient had diabetes.}

Urine chloride is a test that measures the amount of chloride in the urine. Urine chloride is the most appropriate test in this case because the patient's urine chloride is low. Low urine chloride can be caused by a number of conditions, including vomiting, diarrhea, and diuretic use. In this case, the patient's low urine chloride is likely caused by her recent history of nausea, vomiting, and diarrhea.

Urine toxicology is a test that measures the levels of drugs and other toxins in the urine. Urine toxicology is not the most appropriate test in this case because the patient does not have a history of drug abuse and her presentation is not consistent with drug intoxication.
\\
\textbf{RELEVANCE SCORE:} 4 (Mostly relevant)
\\
\bottomrule
	\end{tabular}
	\caption{An example of human annotation about \textbf{relevance}. The highlighted sentence ``\textit{Serum glucose would be more helpful if the patient had diabetes}'' is identified as irrelevant to the question by the annotator.}
	\label{tab:annotation_relevance}
\end{table*}

\end{document}